%% file: camera-ready.tex
\documentclass[runningheads]{llncs}

\usepackage{eccv}

\usepackage{eccvabbrv}
\usepackage{algorithmic}
\usepackage{graphicx}
\usepackage{booktabs}

\usepackage[accsupp]{axessibility}  

\usepackage{hyperref}

\usepackage{orcidlink}

\input{content/__pack}

\begin{document}

\title{Pre-trained Visual Dynamics Representations for Efficient Policy Learning} 

\titlerunning{Pre-trained Visual Dynamics Representations}

\author{Hao Luo\inst{1}\and
Bohan Zhou\inst{1}\and
Zongqing Lu\inst{1,2}\thanks{Corresponding author.}}

\authorrunning{H.~Luo et al.}

\institute{School of Computer Science, Peking University \and Beijing Academy of Artificial Intelligence \\ 
\email{\{lh2000,zongqing.lu\}@pku.edu.cn \quad zhoubh@stu.pku.edu.cn}}
\maketitle

\input{content/00_abstract}

\input{content/01_introduction}

\input{content/02_related}

\input{content/03_preliminaries}

\input{content/04_methodology}

\input{content/05_experiement}

\input{content/06_conclusion}

\clearpage  

\section*{Acknowledgements} 
This work was supported by NSFC under grant 62250068.
%
%
\bibliographystyle{splncs04}
\bibliography{all_ref}

\clearpage
\appendix

\input{content/App_Implementation}
\input{content/App_Samples}

\input{content/App_Extended}


\end{document}

%% file: content/__pack.tex
\usepackage{amsmath}
\usepackage{amssymb}
\usepackage{mathtools}
\usepackage[textsize=tiny]{todonotes}
\usepackage{graphicx}
\usepackage{float} 
\usepackage{bm}
\usepackage{enumerate}
\usepackage{enumitem}
\usepackage{multirow}
\usepackage{adjustbox}
\usepackage{threeparttable}
\usepackage[T1]{fontenc}
\usepackage{algorithmic}
\usepackage{algorithm}
\usepackage{wrapfig}

%% file: content/00_abstract.tex
\begin{abstract}

Pre-training for Reinforcement Learning (RL) with purely video data is a valuable yet challenging problem. Although in-the-wild videos are readily available and inhere a vast amount of prior world knowledge, the absence of action annotations and the common domain gap with downstream tasks hinder utilizing videos for RL pre-training. To address the challenge of \textit{pre-training with videos}, we propose \textbf{P}re-trained \textbf{V}isual \textbf{D}ynamics \textbf{R}epresentations (PVDR) to bridge the domain gap between videos and downstream tasks for efficient policy learning. By adopting video prediction as a pre-training task, we use a Transformer-based Conditional Variational Autoencoder (CVAE) to learn visual dynamics representations. The pre-trained visual dynamics representations capture the visual dynamics prior knowledge in the videos. This abstract prior knowledge can be readily adapted to downstream tasks and aligned with executable actions through online adaptation. We conduct experiments on a series of robotics visual control tasks and verify that PVDR is an effective form for pre-training with videos to promote policy learning.

\keywords{Pre-training with videos \and Reinforcement learning}
\end{abstract}

%% file: content/01_introduction.tex
\section{Introduction}

Recent years have seen groundbreaking advancements in Computer Vision (CV) and Natural Language Processing (NLP), notably enhanced by self-supervised pre-training~\cite{brown2020language,chen2020simple,radford2021learning,dosovitskiy2021an,he2022masked}, showcasing the efficacy of pre-training in distilling the prior knowledge about the world from vast data. As acknowledged few-shot learners~\cite{brown2020language}, pre-trained models have also been applied in Reinforcement Learning (RL)~\cite{sekar2020planning}. Pre-training may improve the poor sample efficiency and generalization ability of online RL, mitigating the demand for extensive costly online interactions, especially in complex visual control tasks~\cite{shridhar2023perceiver}. Compared to the readily available data in vision and natural language pre-training, the collection of data poses a challenge on RL pre-training. Early works use expert demonstrations~\cite{bojarski2016end}, sub-optimal data~\cite{kumar2020conservative}, and reward-free data~\cite{yu2022leverage} for offline policy learning, necessitating specific collection efforts. Considering this, some researches~\cite{torabi2018behavioral,baker2022video,nair2023r3m} turn to more accessible and affordable video data. How in-the-wild videos can be used for RL pre-training 
becomes a promising and valuable topic to explore, which we term \textit{\textbf{pre-traininig with videos}}. 

Pre-training with videos presents both potential and challenges. As an accessible and affordable data source, in-the-wild videos offer a rich repository of dynamic prior knowledge. Through self-supervised learning with video data, models can assimilate prior world knowledge pivotal for effective policy learning, including subject consistency, actions, and transitions~\cite{menapace2021playable,yan2023temporally,bruce2024genie}. Conversely, pre-training with the non-specifically collected videos encounters two main challenges. First, the lack of action annotations hinders the direct integration of videos in policy learning. Second, the common domain gap between videos and downstream tasks, given their content-related but not strictly identical nature, necessitates an online adaptation mechanism. 
Faced with these hurdles, existing methods generally fall into two pathways. Some works merely pre-train visual state representations with frames or temporal sequences from videos~\cite{sermanet2018time,yang2021representation,ma2022vip,seo2022reinforcement,radosavovic2023real,nair2023r3m}. The pre-trained representations benefit the understanding of visual states but inadequately leverage task-relevant dynamics knowledge. The other works view videos as a form of referential demonstration. These works pre-train an intrinsic reward function gauging task progress~\cite{nair2023r3m,zhou2023learning,bruce2023learning} or train an inverse dynamics model for action annotation~\cite{baker2022video,ye2022become}, thereby enabling the imitation of videos. Nonetheless, adapting such models for imitation to domain-gap downstream tasks remains challenging. From this, we question \textit{whether there is a more appropriate form of utilizing videos that harnesses the inherent dynamics knowledge and can also be easily adapted to domain-gap tasks}.

Intuitively, we posit that leveraging more abstract visual dynamics priors from videos holds greater promise for online adaptation. Generically collected videos are abundant in visual dynamics prior knowledge conducive to downstream tasks. For instance, human videos may exhibit continuous and purpose-driven behaviors, and robotic videos may show steady and goal-oriented posture and motion. These abstract visual dynamics priors, encapsulating common knowledge and the essence of motion, offer broad generalization capabilities, surpassing the inherent actions intended for imitation. In response to this insight, we propose \textbf{P}re-trained \textbf{V}isual \textbf{D}ynamics \textbf{R}epresentations (\textbf{PVDR}) for efficient policy learning.  In PVDR, the visual dynamics representations carrying visual dynamics prior knowledge are pre-trained via a video prediction task and online adapted to downstream tasks for policy learning.
 


For PVDR learning, we design a Transformer-based~\cite{vaswani2017attention} visual dynamics model employing a conditional variational autoencoder (CVAE)~\cite{sohn2015learning} to learn the visual dynamics representations. In the pre-training stage, the self-supervised video prediction task facilitates the latent variable of CVAE to effectively represent visual dynamics. We view the latent variables in the dynamics model as our visual dynamics representations. During inference, a batch of visual dynamics representations can be sampled from the learned prior to generate visual plans to choose from. During the online adaptation, we fine-tune the visual dynamics model with online experiences and train an action alignment module to turn the chosen plans into executable actions. For valid action alignment, we employ supervised learning with online experiences and RL with a reward signal related to the consistency between the chosen plan and the visual observations during action execution. Through this two-stage pipeline, the PVDR algorithm can learn visual dynamics representations from the pre-training videos and conduct planning-based inference in the downstream tasks, where the agent learns to turn plans into actions with feedback from the environment.

\begin{wrapfigure}[15]{r}{0.5\textwidth}
\setlength{\abovecaptionskip}{2mm}
    \centering
    \includegraphics[width=\linewidth]{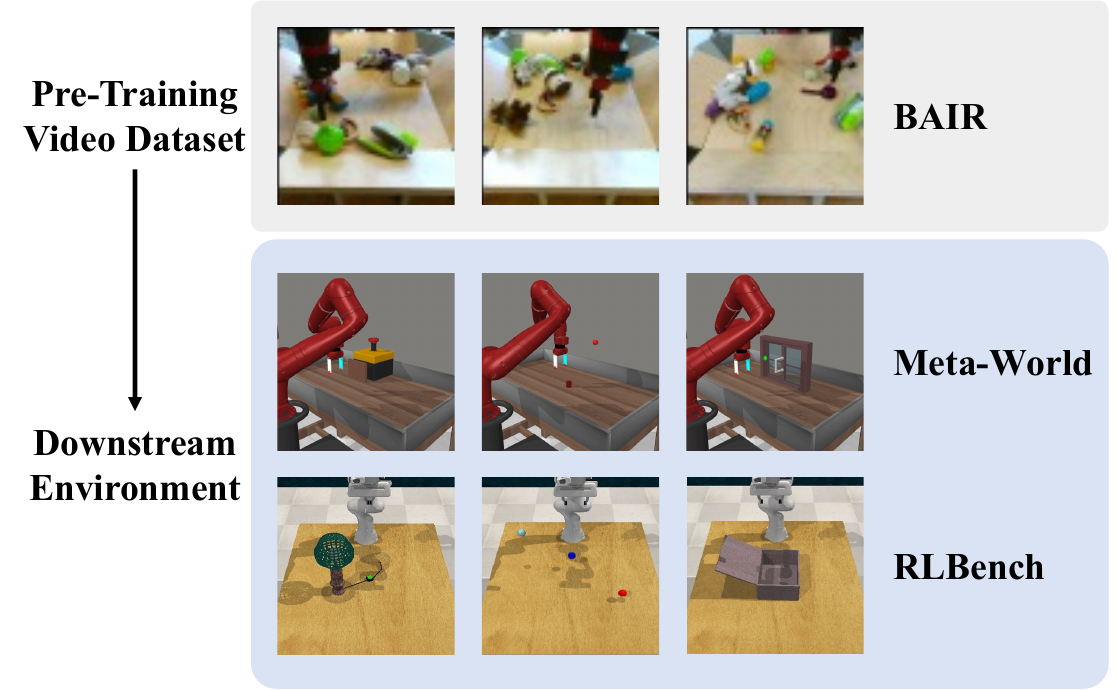}
    \caption{Illustration of the pre-training video dataset and the downstream environments we used for experimental evaluations.}
    \label{fig:env}
\end{wrapfigure}

To evaluate the proposed PVDR algorithm, we use the BAIR robot pushing dataset~\cite{ebert2017self} recording the real-world robotic arm as the pre-training video dataset and test PVDR on goal-conditioned control tasks in the Meta-World~\cite{yu2020meta} and RLBench~\cite{james2020rlbench} environments, which is illustrated in Figure~\ref{fig:env}. 
In the experiments, PVDR outperforms baselines, and ablation studies verify the effectiveness of the core design. Our experimental results indicate that visual dynamics representations are a suitable form to bridge domain-gap videos and downstream visual control tasks, and our online adaptation is conducive to turning visual dynamics representations into aligned actions. Thus, PVDR is an effective algorithm to potentially address the challenge of {pre-trainning with videos}.

%% file: content/02_related.tex
\section{Related Work}

Many previous works have studied leveraging purely video data to aid policy learning. In this setting, different works utilize the prior information in videos from different perspectives and to varying degrees.

\noindent\textbf{State Representations Pre-trained with Videos.} Considering the lack of action annotations in the video data, some works pre-train visual state representations to aid in understanding high-dimensional and complex visual observations. The compressed state representation space allows agents to learn policies more efficiently in downstream tasks. These works utilize video data through methods such as masked auto-encoder~\cite{xiao2022masked,radosavovic2023real,yang2024spatiotemporal}, time-contrastive learning~\cite{sermanet2018time,nair2023r3m,li2024auxiliary}, value function learning~\cite{ma2022vip,bhateja2023robotic,ghosh2023reinforcement}, and dynamics learning~\cite{seo2022reinforcement,wu2023pre}. To some extent, the pre-trained state representations leverage the dynamics knowledge in videos. However, the absence of an explicit mechanism to utilize such knowledge limits their promotion of policy learning.

\noindent\textbf{Imitation-Intended Pre-training with Videos.} Other works view videos as action-free demonstrations, using video pre-training to guide agent to imitate the videos. To address the lack of action annotations, different techniques have been incorporated. Some straightforward methods involve an Inverse Dynamics Model (IDM)~\cite{pathak2018zero,schmeckpeper2021reinforcement,baker2022video,zhang2022learning,zheng2023semi} or a latent IDM~\cite{edwards2019imitating,schmeckpeper2021reinforcement,ye2022become,schmidt2023learning} to label actions or latent actions in videos. 
However, when there is a gap in the action space between pre-training videos and downstream tasks, (latent) IDMs are likely to fail. Unlike IDM-based methods, some methods learn reward functions from videos to assess the degree of task completion, followed by online RL with the learned intrinsic reward functions. The reward functions are relevant to various unsupervised signals, such as the log-likelihood of predicted videos~\cite{escontrela2024video}, the divergence to generated transitions~\cite{yu2020intrinsic}, the temporal distance~\cite{bruce2023learning}, the potential promotion of state value~\cite{chang2021learning,bobrin2024align}, and the transition discrimination scores~\cite{torabi2018behavioral,torabi2018generative,torabi2019imitation,yang2019imitation,zhou2023learning}. These reward functions encourage the imitation of offline videos from different perspectives. Similarly, some works~\cite{nair2020contextual,sharma2019third} learn a (sub-)goal generation module to guide visual imitation. Although generalization has been observed in some composite scenarios, these imitation-intended methods still suffer in tasks with dissimilar intentions~\cite{escontrela2024video}. Furthermore, some works~\cite{peng2018sfv,bahl2022human,mendonca2023structured,bahl2023affordances} integrate video feature extraction models with designed priors, such as motion extraction, pose estimation, and affordance prediction, and guide policy learning with extracted features. However, these methods are task-specific thus not general.

\noindent Compared to previous works, our work employs visual dynamics representations pre-trained from videos. Unlike state representations, visual dynamics representations explicitly encode the dynamics in videos into one vector, thus enabling the alignment with actions in the online stage and the direct participation in policy learning. In contrast to imitation-intended methods, the dynamics information of video snippets is in a more high-level and abstract form. Such visual dynamics prior knowledge can easily adapt in the downstream tasks, rather than imitating the state transitions in the pre-training videos.

Among the previous works, APV~\cite{seo2022reinforcement} and FICC~\cite{ye2022become} are the most closely related to our work, both utilizing pre-trained dynamics models to aid online policy learning with visual dynamics knowledge. APV pre-trains an observation-only dynamics model to understand the visual dynamics in videos. However, it employs the dynamics information solely as an observation representation rather than as a high-level dynamics prior, which limits its effectiveness. FICC pre-trains a forward-inverse dynamics model to learn a latent action from two adjacent frames. To perform a Monte Carlo Tree Search (MCTS) in the environment, FICC additionally learns a mapping from the latent actions to the real actions. However, when there is a gap in the action space between pre-training videos and downstream tasks, such mapping is hard to learn and work with.

More related works are discussed in Appendix~\ref{app:extended}.

%% file: content/03_preliminaries.tex
\section{Preliminaries}

\subsection{Goal-Conditioned POMDP}
\label{sec:GCPOMDP}
In this work, we consider goal-conditioned visual control tasks, employing a goal-conditioned variant of the partially observable Markov decision process (POMDP)~\cite{aastrom1965optimal,kaelbling1998planning}. A goal-conditioned POMDP is represented by a seven-tuple $\left(\mathcal{S}, \mathcal{A}, \mathcal{T}, g, \mathcal{O}, \Omega, \gamma\right)$, where $\mathcal{S}, \mathcal{A}, g, \mathcal{O}, \gamma$ respectively denote the state space, the action space, the image goal, the observation space and the discount factor. For each task, the visual observation of the target state is provided as the goal $g$. At each timestep $t$, the agent receives an observation $o_t \in \mathcal{O}$, which is mapped from the current state $s_t \in \mathcal{S}$ by the observation function $\Omega: \mathcal{S} \mapsto \mathcal{O}$. After the agent takes an action $a_t \in \mathcal{A}$, the environment transitions to the next state $s_{t+1}$ according to the transition function $\mathcal{T}: \mathcal{S} \times \mathcal{A} \times \mathcal{S} \mapsto [0,1]$. The agent's objective is to reach the state indicated by the task's goal. 

As our work does not focus on the reward function, we simply assign $r_t = - \left\|o_{t+1} - g\right\|_1$ as a reasonable reward function, which is a common and straightforward form in goal-conditioned RL~\cite{ebert2017self,wu2018laplacian,pong2020skew}.

\subsection{Video Prediction}

In addressing the challenge of pre-training with videos, we utilize video prediction~\cite{ranzato2014video} as a pre-training task to capture visual dynamics knowledge. Concretely, at timestep $t$ in a video, we formulate the video prediction task by predicting $l$ future frames $\mathbf{Y}_t \coloneqq \left(Y_{t+1}, \cdots, Y_{t+l}\right)$ based on $m$ context frames of the past and present $\mathbf{X}_t \coloneqq \left(X_{t-m+1}, \cdots, X_t\right)$. The video prediction task is to model the distribution $Pr\left(Y_{t+1}, \cdots, Y_{t+l} \mid X_{t-m+1}, \cdots, X_t\right)$, which is indeed a natural form of visual dynamics.

\begin{wrapfigure}[19]{r}{0.42\textwidth}
\setlength{\abovecaptionskip}{2mm}
    \centering
    \includegraphics[width=\linewidth]{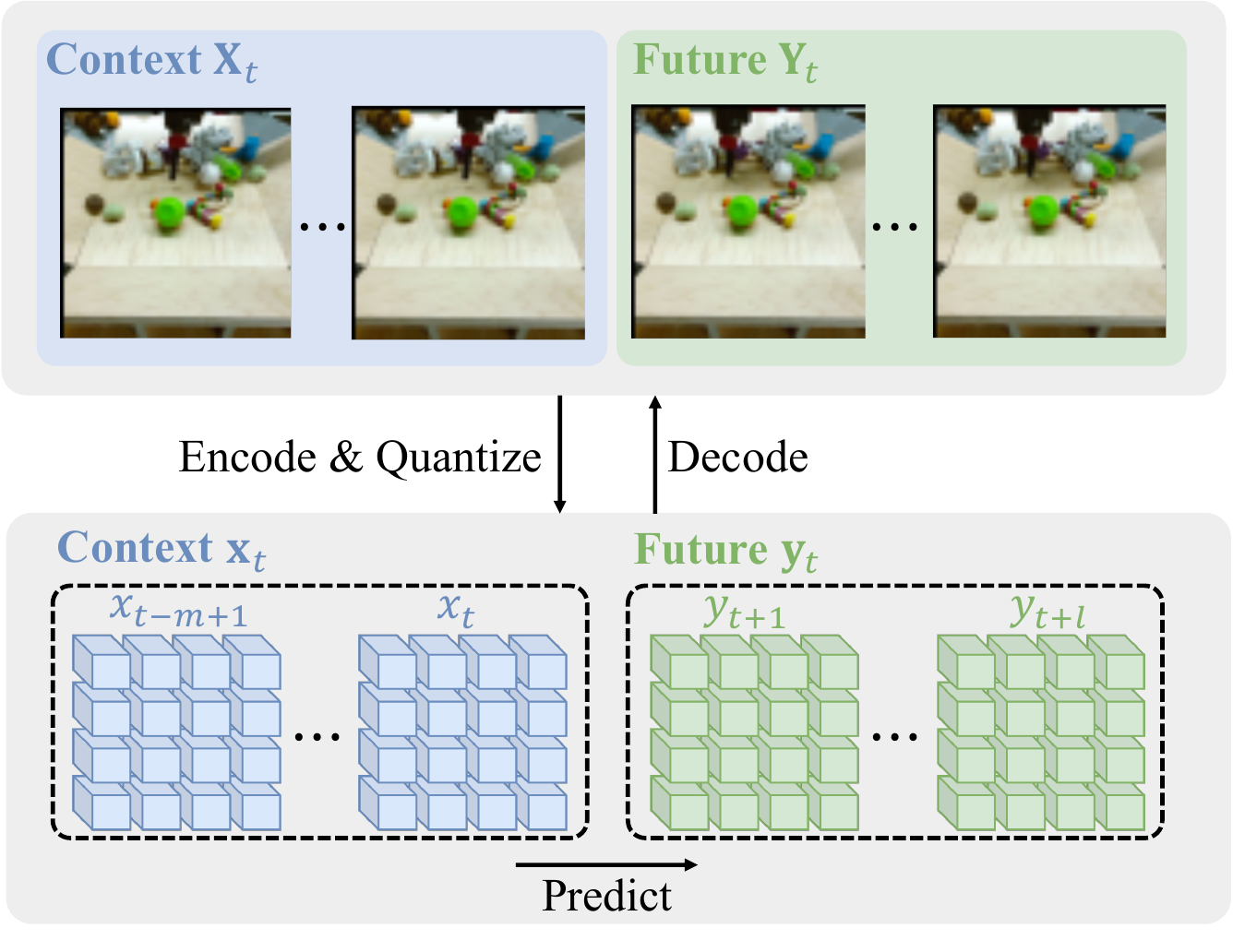}
    \caption{Illustration of video prediction with dVAE. In this paradigm, a dVAE is pre-trained to compress frames from raw pixel-level space (upper block) into discrete latent space (lower block) for effective prediction.}
    \label{fig:vq}
\end{wrapfigure}

Recent developments~\cite{walker2021predicting,gupta2022maskvit,yan2023temporally} in the video prediction task have ushered a paradigm utilizing a pixel-level discrete variational autoencoder (dVAE), \eg VQVAE~\cite{van2017neural} and VQGAN~\cite{esser2021taming}, compressing frames into a smaller grid of visual tokens for better prediction. As illustrated in Figure~\ref{fig:vq}, these methods model an visual distribution autoregressively with a causal Transformer in the compressed latent space. Formally, with a codebook $\mathcal{Q} = \left\{q_{k}\right\}_{k=1}^{K} \subset \mathbb{R}^{n_{q}}$, each frame $X_t \in  \mathbb{R}^{H \times W \times 3}$ is encoded and quantized into a grid of discrete tokens: $x_t \coloneqq {\operatorname{argmin}}_{q \in \mathcal{Q}}\left\| q - \operatorname{Enc}(X_t) \right\|_2 \in \mathbb{R}^{h \times w \times n_q}$. Similarly, $Y_t \in \mathbb{R}^{H \times W \times 3}$ is also mapped to $y_t \in \mathbb{R}^{h \times w \times n_q}$. Compared to prediction in the huge raw pixel space, it is more effective to conduct video prediction via modeling the distribution $Pr\left(y_{t+1}, \cdots, y_{t+l} \mid x_{t-m+1}, \cdots, x_t\right)$ in a compressed discrete latent space.

We adopt this paradigm to effectively address the video prediction task in a compressed discrete latent space. For simplicity in the following discussion, we use the notation $\mathbf{x}_t \coloneqq \left(x_{t-m+1}, \cdots, x_t\right)$ for compressed context frames in the video prediction task and compressed context observations in the decision-making task, correspondingly $\mathbf{y}_t \coloneqq \left(y_{t+1}, \cdots, y_{t+l}\right)$ for compressed future frames/ observations.

%% file: content/04_methodology.tex
\section{Methodology}

\begin{figure}[tb]
\setlength{\abovecaptionskip}{2mm}
    \centering
    \includegraphics[width=\linewidth]{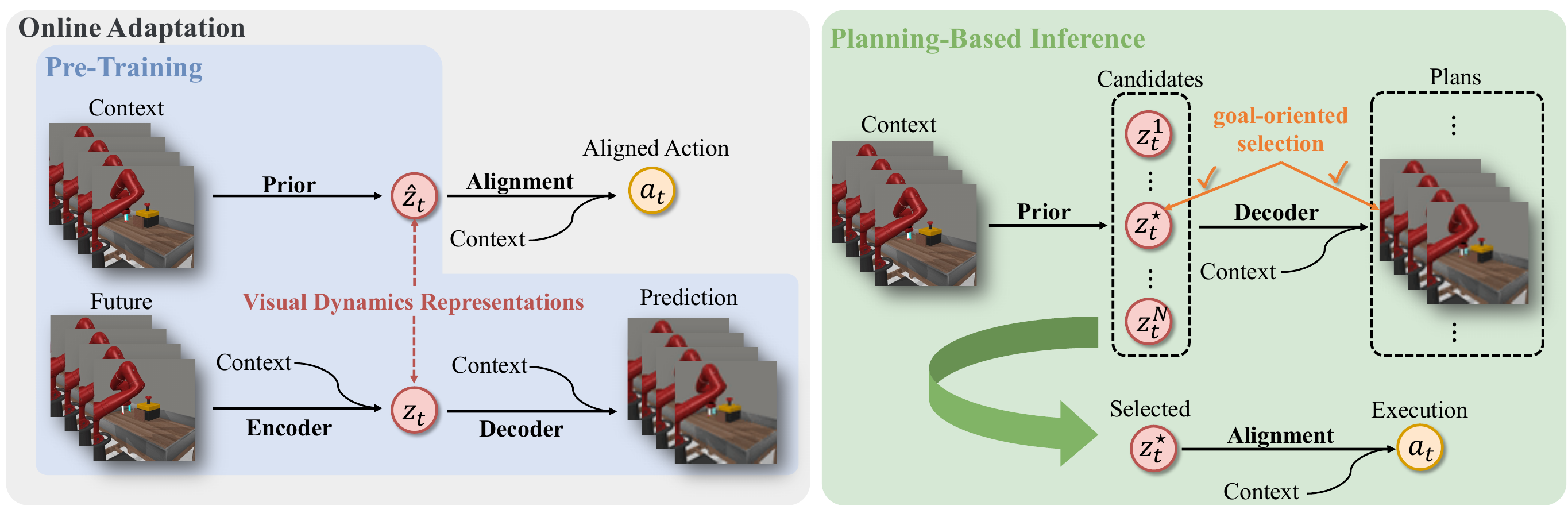}
    \caption{Illustration of PVDR workflow for different stages and usages. Four modules, \textit{Visual Dynamics Encoder}, \textit{Visual Dynamics Decoder}, \textit{Visual Dynamics Prior}, and \textit{Action Alignment Module}, are included in PVDR.  The learning workflow is on the left, while the inference workflow is on the right. In the pre-training stage (blue block), \textit{Encoder}, \textit{Decoder}, and \textit{Prior} are pre-trained to capture visual dynamics representations. During the online adaptation (gray block), the \textit{Action Alignment Module} is integrated with the other three modules. Planning-based inference (green block) is performed to choose the plans closest to the goal and execute aligned actions.}
    \label{fig:workflow}
\end{figure}

In this section, we delineate the comprehensive workflow of PVDR. As a two-stage algorithm, PVDR pre-trains a visual dynamics model with the video prediction task in the pre-training stage and adapts the model to the downstream tasks with an integrated action alignment module in the online stage. 
The overall process is illustrated in Figure~\ref{fig:workflow}.

\subsection{Pre-training Stage}
\label{sec:pre-training}
In the pre-training stage, a visual dynamics model is pre-trained to capture the visual dynamics prior knowledge intrinsic in the videos. Specifically, the visual dynamics information of the visual sequence formed by $ \mathbf{x}_t $ and $ \mathbf{y}_t $ is supposed to be encapsulated as a latent variable $ z_t $. As shown in Figure~\ref{fig:workflow}.(blue), the visual dynamics model employs the conditional variational autoencoder (CVAE). The visual dynamics model is composed of three pivotal components:
\begin{itemize}
    \item[] \textbf{Visual Dynamics Encoder} $ E_\zeta $: It encodes a latent representation $ z_t$ of the video snippet containing context frames $ \mathbf{x}_t $ and grounded future frames $ \mathbf{y}_t $. Formally, $z_t\sim E_\zeta(\cdot \mid \mathbf{x}_t, \mathbf{y}_t)$.
    \item[] \textbf{Visual Dynamics Decoder} $ D_\eta $: It uses context frames $ \mathbf{x}_t $ and the encoded representation $ z_t $ to reconstruct future frames $ \hat{\mathbf{y}}_t $. Formally, $\hat{\mathbf{y}}_t \sim D_\eta(\cdot \mid \mathbf{x}_t, z_t)$.
    \item[] \textbf{Visual Dynamics Prior} $ P_\theta $: It models the distribution of the latent representation $ \hat{z}_t $ given context frames $ \mathbf{x}_t$. Formally, $\hat{z}_t \sim P_\theta(\cdot \mid \mathbf{x}_t) $.
\end{itemize}

The visual dynamics encoder $E_\zeta$ and the visual dynamics decoder $D_\eta$ are optimized to maximize the log-likelihood of the ground truth. In order to accurately predict incoming frames, the optimization forces the encoder to capture the latent representation $z_t$ informative with the visual dynamics. Thus, $z_t$ is indeed a representation of visual dynamics. 

Additionally, a standard Gaussian distribution $\mathcal{N}(\mathbf{0}, \mathbf{I})$ is employed as an information bottleneck to prevent the visual dynamics representations from being overloaded with information. Otherwise, the model may collapse and impair the effective extraction of relevant information. Furthermore, we incorporate a KL divergence loss to learn the prior $P_\theta(\cdot \mid \mathbf{x}_t)$ with the posterior of the visual dynamics encoder $E_\zeta(\cdot \mid \mathbf{x}_t, \mathbf{y}_t)$. Comprehensively, the loss of the pre-training stage is $\mathcal{L}_{pre}$, 
\begin{small}
    \begin{align}
            \mathcal{L}_{pre}(\zeta,\eta,\theta) =& \underbrace{-\log D_\eta \left(\mathbf{y}_t\mid\mathbf{x}_t, z_t \hat{\sim} E_\zeta(\cdot \mid \mathbf{x}_t, \mathbf{y}_t)\right)}_{\text{reconstruction loss}}
    + \lambda_1 \cdot \underbrace{D_{\text{KL}}\left(E_\zeta(\cdot \mid \mathbf{x}_t, \mathbf{y}_t) \|\mathcal{N}(\mathbf{0}, \mathbf{I}) \right)}_{\text{information bottleneck}} \notag \\
    &+ \underbrace{D_{\text{KL}}\left(P_\theta(\cdot \mid \mathbf{x}_t)\| \operatorname{sg} [E_\zeta(\cdot \mid \mathbf{x}_t, \mathbf{y}_t)]\right)}_{\text{prior loss}} ,
    \label{loss:pre}
    \end{align}
\end{small}

\noindent where $\hat{\sim}$ and \textit{$\operatorname{sg}$} respectively refer to the reparameterization operation and the stop-gradient operation, and $\lambda_1$ is a hyperparameter.

Via pre-training a visual dynamics model, $E_\zeta$ encodes the visual dynamics information of video sequences, and $P_\theta$ captures the visual dynamics prior knowledge from the videos. Together, $P_\theta$ and $D_\eta$ can perform video prediction based on context frames. All of these elements are utilized during the online stage.

\subsection{Online Stage: Inference}
\label{sec:inference}
The online stage consists of two alternating and repeating parts: inference with the pre-trained visual dynamics model for experience collection, and adaptation to the downstream tasks with the collected experience. For ease of description, we first describe how the learned dynamics model is utilized during inference, before discussing adaptation in Section~\ref{sec:align}. Thus, we assume that the model has already been adapted to a certain downstream task in this subsection.

The inference process is designed as a planning paradigm anchored with the pre-trained visual dynamics representations. Intuitively, given context observations and various visual dynamics representations sampled from the prior, the visual dynamics model can generate a batch of possible predictions as visual plans. From the sampled representation candidates, the one with the closest plan to the goal is selected as the desired visual dynamics representation. In addition, an action alignment module $\Pi_\phi$ is integrated to transform the selected visual dynamics representation into tangible actions. The action alignment module $\Pi_\phi$ is tailored to generate action given context frames and the chosen visual dynamics representation, thereby actualizing the visual dynamics plan envisaged. How to learn such an action alignment module is discussed in Section~\ref{sec:align}.

As illustrated in Figure~\ref{fig:workflow}.(green), given the context observations $\mathbf{x}_t$, a batch of visual dynamics representation candidates are generated from the prior $P_\theta$. A desired visual dynamics representation $z^\star_t$ is chosen according to:
\begin{align}
    z^\star_t \coloneqq \underset{z_t\in \mathcal{Z}_t }{\operatorname{argmin}} \left\|D_\eta(\mathbf{x}_t,z_t) - g\right\|_1 , 
\end{align}
where $\mathcal{Z}_t \coloneqq \left\{z_t^n\sim P_\theta(\cdot \mid \mathbf{x}_t)\right\}^N_{n=1}$ refers to the set of visual dynamics representation candidates.  Then an action $a_t$ is generated as $a_t \sim \Pi_\phi ( \cdot \mid z^\star_t, \mathbf{x}_t)$ and executed.  \textit{The whole inference process of first planning and then aligning $z^\star_t$ with the specific action enables the agent to interact with the downstream environment.}

\subsection{Online Stage: Model Adaptation and Action Alignment}
\label{sec:align}
In the planning-based inference process described in Section~\ref{sec:inference}, the pre-trained model is supposed to adapt to the downstream tasks with the collected experiences. Concurrently, an action alignment module will learn from scratch to align with the desired visual dynamics representations. Three aspects should be taken into consideration for adaptation and alignment: (1) fine-tuning the dynamics model to generate visual sequences following the visual distribution of downstream tasks; (2) fine-tuning the dynamics prior to generate better visual dynamics representation candidates to reach the goal more easily than the general ones from the pre-training dataset; (3) aligning the visual dynamics representations with reasonable actions via the action alignment module.
 The whole process of online adaptation and alignment is shown in Figure~\ref{fig:workflow}.(gray). 

\noindent\textbf{Experience.} First, we denote the form of experiences utilized in online adaptation and alignment. Throughout the interaction with the environment, a quintuple $e_t \coloneqq \left(\mathbf{x}_t, \mathbf{y}_t, z^{\star}_t, \hat{\mathbf{y}}^\star_t, a_t\right)$ is recorded at each timestep $t$. Here, $\mathbf{x}_t, \mathbf{y}_t$ represent the context and future observations at timestep $t$ in the entire trajectory. Moreover, $z^{\star}_t, \hat{\mathbf{y}}_t, a_t$ respectively denote the visual dynamics representation selected via planning, its corresponding visual plan, and the aligned action from the action alignment module. Each trajectory is recorded as a set of such quintuples $\mathcal{E} \coloneqq \left\{e_t\right\}_{t=0}^{T-1}$.

\noindent\textbf{Visual Dynamics Fine-tuning.} The visual dynamics model is fine-tuned to adapt to the visual distribution of the downstream environment. Using $\mathbf{x}_t, \mathbf{y}_t$ from the experience, we fine-tune the dynamics encoder and decoder with the reconstruction loss term consistent with the pre-training stage\footnote{To maintain the pre-trained visual dynamics knowledge, we freeze the dynamics encoder except for the last layer that generates the mean and variance of $z_t$.},
\begin{align}
    \mathcal{L}_{rec}(\eta,\zeta)=-\log D_\eta \left(\mathbf{y}_t\mid\mathbf{x}_t, z_t \hat{\sim} E_\zeta(\cdot \mid \mathbf{x}_t, \mathbf{y}_t)\right).
    \label{loss:rec}
\end{align}
The visual dynamics prior is also fine-tuned to adapt to the downstream distribution with the prior loss term in the pre-training stage. Besides, we tend to adapt the prior  $P_\theta$ to generate visual dynamics representations more conducive to reaching the goal of the task. To achieve this, we use the distance between the goal and the predicted plans as a loss to encourage the generation of visual dynamics representation candidates closer to the goal. Formally, we utilize the following loss to fine-tune the prior $P_\theta$,
\begin{small}
\begin{align}
    \mathcal{L}_{prior}(\theta) =& \underbrace{D_{\text{KL}}\left(P_\theta(\cdot \mid \mathbf{x}_t)\| \operatorname{sg} [E_\zeta(\cdot \mid \mathbf{x}_t, \mathbf{y}_t)]\right)}_{\text{prior loss}} 
    +\lambda_2\cdot\underbrace{\| \hat{y}_t \sim D_\eta(\cdot|\mathbf{x}_t,z_t\hat{\sim}P_\theta(\cdot|\mathbf{x}_t))  - g\|_1}_{\text{goal-oriented loss}},
    \label{loss:prior}
\end{align}    
\end{small}
where $\lambda_2$ is a hyperparameter.



\noindent\textbf{Action Alignment Learning.} 
The most crucial problem of the online stage is learning to align the visual dynamics representations with actions. Action alignment is to find the corresponding actions that turn the envisioned visual dynamics plan into reality. We use the online interaction experience $\left(\mathbf{x}_t, \mathbf{y}_t, z^{\star}_t, \hat{\mathbf{y}}_t, a_t\right)$ to learn this alignment. It is evident that the visual dynamics of $(\mathbf{x}_t, \mathbf{y}_t)$ can be achieved by executing action $a_t$ under the context observations $\mathbf{x}_t$. Therefore, we can use the visual dynamics encoder to capture the visual dynamics representation $z_t$ of $(\mathbf{x}_t, \mathbf{y}_t)$, and $z_t$ under $\mathbf{x}_t$ should be aligned with the action $a_t$. 
In this way, we can obtain data pairs of visual dynamics representations and their corresponding actions from the experience and use such data pairs for supervised learning. Formally, given the encoded visual dynamics representation $z_t \sim E_\zeta(\cdot \mid \mathbf{x}_t,\mathbf{y}_t)$, the loss is 
\begin{align}
    \mathcal{L}_{act}(\phi) &= \|a_t -  \hat{a}_t \hat{\sim} \Pi_\phi(\cdot \mid z_t, \mathbf{x}_t) \|_2.
    \label{loss:act}
\end{align}

However, relying solely on supervised learning from observations and actions in the experiences has limitations. Before the policy converges, the visual dynamics $(\mathbf{x}_t, \mathbf{y}_t)$ in the experiences are evidently far from the goal. Thus, such a visual dynamics representation $z_t$ is unlikely to be selected by planning. In other words, there is a discrepancy between the distribution of the visual dynamics representation $z_t$ of the actual experience and the visual dynamics representation $z^{\star}_t$ chosen during planning. That is, the planning-selected $z^{\star}_t$ is hard to be covered by the experienced $z_t$ before extensive interactive exploration, hindering efficient visual plan execution. 
Since the ground truth of the action corresponding to $z^{\star}_t$ is unavailable, we additionally design a reward, which encourages $\Pi_\phi$ to generate a reasonable action aligned with $z^{\star}_t$, and optimize the action alignment module using RL based on this reward. We define the rationality of an action from two perspectives. From the perspective of goal completion, the action should facilitate the transition to the next observation $y_{t+1}$ towards the visual goal. From the perspective of plan adherence, the action should make the future observations $\mathbf{y}_t$ in the environment closer to the visual plan $\mathbf{\hat{y}}_t$. Specifically, we use a reward of the form:
\begin{align}
    \tilde{r}_t = \underbrace{\|y_{t+1} - g \|_1}_{\text{goal-oriented}}+ \lambda_3 \cdot \underbrace{\frac{1}{l}\|\mathbf{y}_t-\mathbf{\hat{y}}_t\|_1}_{\text{planning-oriented}},
    \label{eq:r_label}
\end{align}
where $l$ is the frame number of $\mathbf{y}_t$ used for normalization and $\lambda_3$ is a hyperparameter. Viewing $(\mathbf{x}_t, z^{\star}_t)$ as the state and $a_t, \tilde{r}_t$, respectively, as the action and the reward, a new MDP is constructed. In this MDP, the action alignment function $\Pi_\phi$ outputs action conditioned on the state. Thus, $\Pi_\phi$ acts as a policy, which can be optimized using RL algorithms such as PPO~\cite{schulman2017proximal}. When the cumulative reward is greater, the generated action leads the agent to get close to the goal and close to the visual plan. Optimized with supervised learning and RL jointly, the action alignment module learns to align the visual dynamics representations with reasonable actions.

\subsection{Overall Procedure}
\label{sec:overall}
PVDR combines a visual dynamics model with an action alignment module to form a planning-based policy that leverages pre-trained visual dynamics representations. PVDR is a two-stage method, which is outlined in Algorithm~\ref{alg:PVDR}. It is worth noting that during the online adaptation stage, we separately conduct the fine-tuning of the dynamics model and the learning of the action alignment module. This separation is to mitigate the optimization difficulty arising from the strong correlation between their optimization objectives.  

    \begin{algorithm}[tb]
       \caption{PVDR}
       \label{alg:PVDR}
    \begin{algorithmic}[1]
        \STATE {\textbf{Require: }} pre-training video dataset $\mathcal{D}$.
       \STATE {\textbf{Initialize: }} visual dynamics encoder $E_\zeta$, visual dynamics decoder $D_\eta$, visual dynamics prior $P_\theta$, action alignment module $\Pi_\phi$.
       \REPEAT
        \STATE Optimize $E_\zeta, D_\eta, P_\theta$ with $\mathcal{L}_{pre}$ in \Cref{loss:pre} on $\mathcal{D}$. 
       \UNTIL{modules are converged.}{ // \textit{\textbf{pre-training stage}}}
        \REPEAT
        
        \STATE Collect experiences buffer $\mathcal{E}_1$ through planning.
        \STATE Optimize $E_\zeta, D_\eta$ with $\mathcal{L}_{rec}$ in \Cref{loss:rec} on $\mathcal{E}_1$.
        \STATE Optimize $P_\theta$ with $\mathcal{L}_{prior}$ in \Cref{loss:prior} on $\mathcal{E}_1$.
        
        \STATE Collect experiences buffer $\mathcal{E}_2$ through planning.
        \STATE Label each piece of experience in $\mathcal{E}_2$ with $\tilde{r}_t$ in \Cref{eq:r_label} .
        \STATE Optimize $\Pi_\phi$ with $\mathcal{L}_{act}$ in \Cref{loss:act} and PPO loss on $\mathcal{E}_2$.
        \UNTIL{modules are converged.}{ // \textit{\textbf{online stage}}}
    \end{algorithmic}
    \end{algorithm}

%% file: content/05_experiement.tex
\section{Experimental Results}

To validate the effectiveness of PVDR, we use the BAIR robot pushing dataset~\cite{ebert2017self} for pre-training and then fine-tune and test the model on a series of tasks in Meta-World~\cite{yu2020meta} and RLBench~\cite{james2020rlbench}. The BAIR robot pushing dataset comprises videos in which a Sawyer robot pushes a variety of objects on a table. The dataset is widely used for video prediction and contains the visual dynamics prior conducive to downstream visual control tasks. However, there is a domain gap between the videos and the downstream environments, as shown in Figure~\ref{fig:env}. We design our experiments to answer the following questions:
\begin{itemize}
    \item[1.] Whether visual dynamics representations are effective for utilizing domain-gap videos in downstream visual control tasks.
    \item[2.] The effectiveness of various optimization objectives in the online adaptation stage of the PVDR algorithm.
    \item[3.] Whether visual dynamics representations ultimately capture action-aligned information.
\end{itemize}

\subsection{Experiments Setup}

\textbf{Environments.} The Meta-World and RLBench environments are widely adopted for validation and provide a variety of challenging robot visual control tasks. In our experiments, we choose 12 tasks from Meta-World and 3 tasks from RLBench. In all Meta-World tasks, the episode length is 500 steps with action repeat of 2 and we use the customized \texttt{camera1} to capture observations. In all RLBench tasks, the episode length is 200 steps with action repeat of 2 and we use the cropped observations from the front camera. In all pre-training videos and downstream tasks, we resize the frames/observations to $128\times128$. 
As we consider goal-based tasks, we replace the original dense reward in Meta-World with the sparse reward, and a positive $r_{suc}$ will be given when the goal is achieved. Additionally, we fix the initial pose of objects in RLBench tasks for definite visual goals.

\noindent\textbf{Baselines.} We compare PVDR with three baselines addressing pre-training with videos, including:
\begin{itemize}
    \item[]  \textbf{APV}\cite{seo2022reinforcement} pre-trains observation representations via auto-regression dynamics prediction. In downstream tasks, a Dreamer-style~\cite{hafner2021mastering} paradigm is built on the observation representations.
    \item[] \textbf{FICC}\cite{ye2022become} pre-trains a latent action space with a forward-inverse dynamics model. In downstream tasks, the pre-trained dynamics model is used for an MCTS-style algorithm, with an online-learned action mapping.
    \item[] \textbf{STG}~\cite{zhou2023learning} pre-trains an intrinsic reward function to encourage the imitation of videos. The reward function is used for PPO in downstream tasks.
\end{itemize}
In addition, we include \textbf{PPO} and \textbf{PVDR without pre-training} as the baselines for without pre-training.

\noindent\textbf{Implementation.} In consistent with some recent works in the video prediction realm, all modules\footnote{Actor-critic algorithms require a critic network, and we simply replace the last layer of the action alignment module to construct a critic network.} in PVDR are Transformer-based networks. Additionally, to alleviate the computational burden of attention, we adopt the Bidirectional Window Transformer in MaskViT~\cite{gupta2022maskvit}. 
In all tasks, we run each algorithm five times and report the average success rate along with the standard deviation. More details of the implementation and hyperparameters are available in Appendix~\ref{app:experiments}.

\subsection{Meta-World Experiments}

\begin{figure}[tb]
    \centering
    \includegraphics[width=0.95\linewidth]{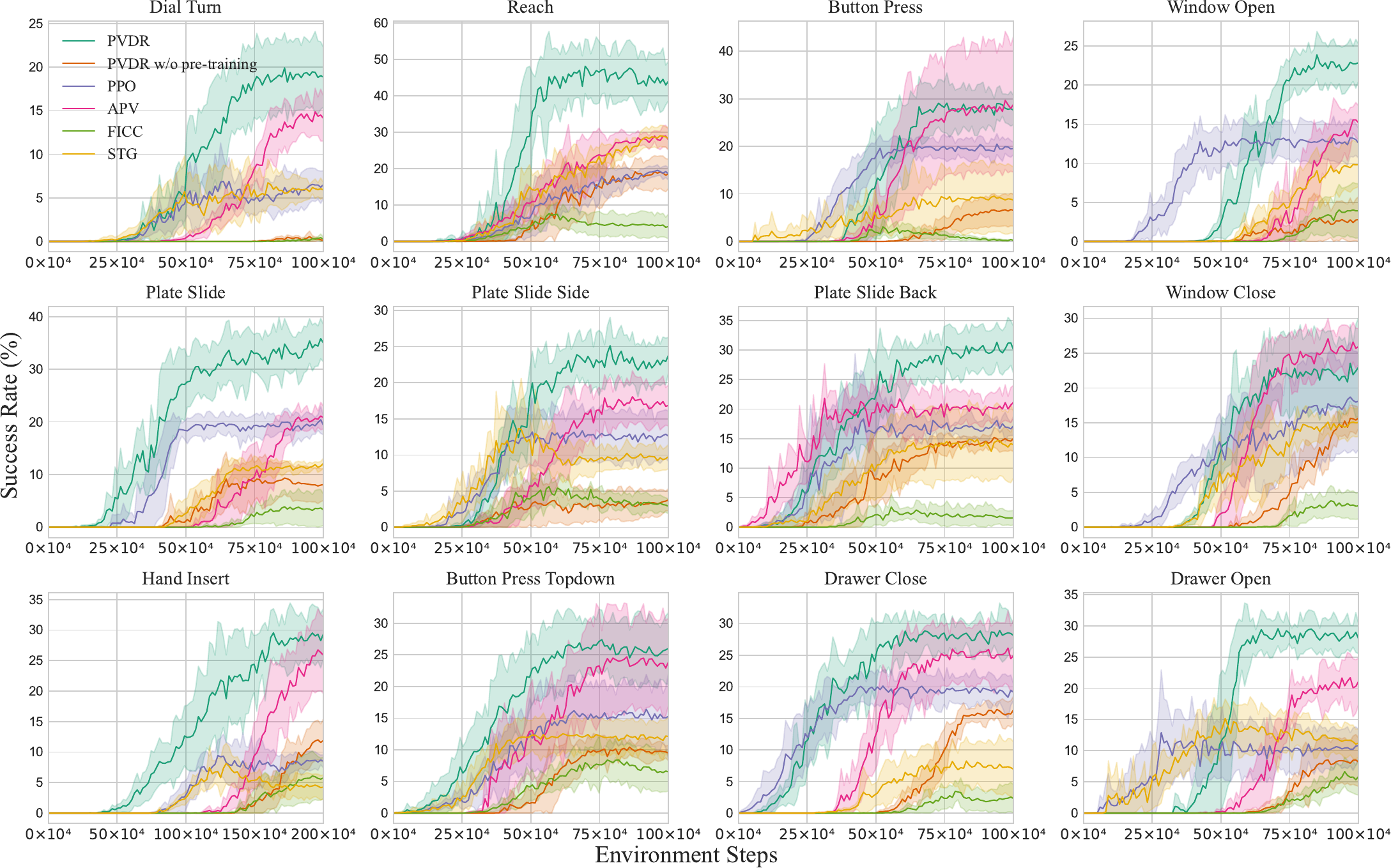}
    \caption{Learning curves of PVDR compared with five other baselines on 12 Meta-World tasks measured on success rate. The solid line and shaded regions represent the mean and variance of the performance across five runs with different seeds.}
    \label{fig:mw.baseline}
\end{figure}

\noindent\textbf{Comparison with Baselines.}
The learning curves of PVDR and baselines on Meta-World are shown in Figure~\ref{fig:mw.baseline}, recording the success rate throughout the course of online interaction steps. Overall, PVRD outperforms the baselines and obtains the best performance in 7 out of 12 tasks. APV obtains comparable performance with PVDR in the rest of 5 tasks but exhibits lower sample efficiency in general. 
 The effectiveness of visual dynamics representations is validated via comparison with the performance of baselines from several perspectives:
\begin{itemize}
    \item The advantage of PVDR over purely online baselines, PPO and PVDR without pre-training, suggests that visual dynamics representations effectively capture knowledge from the pre-training dataset and subsequently benefit the downstream tasks.
    \item Regarding STG, the reliance on in-domain demonstrations leads to the absence of a fine-tuning mechanism with downstream online experiences. Visual dynamic representations may be a better choice when there is a visual gap between pre-training data and downstream scenarios. The performance gap between STG and PVDR also stands with this point.
    \item Although both FICC and PVDR are foresight methods, PVDR shows a notable edge. The advantage might stem from encoding the more abstract and compressed information from a longer horizon, rather than discrete latent action extraction from two adjacent frames in FICC. Therefore, visual dynamics representations might be more suitable for scenarios where pre-training videos can provide behavioral insights for downstream tasks but have less direct correlation in the action space.  
    \item Both PVDR and APV train representations on longer video sequences during pre-training. However, APV pre-trains an observation representation for each frame. Thus, there is no explicit visual dynamics prior in APV, which might be a reason for the better performance of PVDR.
\end{itemize}

\begin{figure}[tb]
    \centering
    \includegraphics[width=0.95\linewidth]{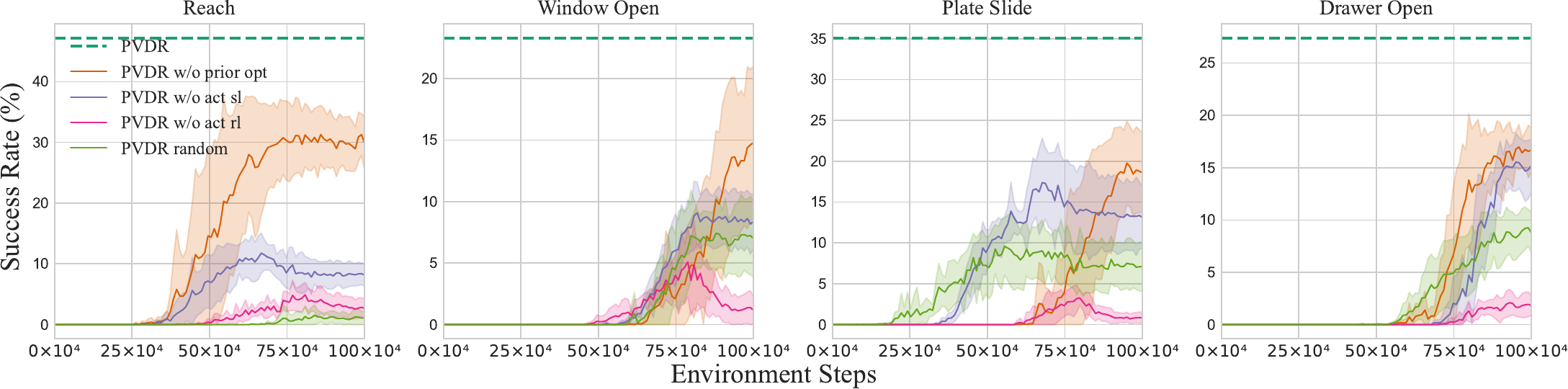}
    \caption{Learning curves of PVDR's ablation studies on 4 Meta-World tasks measured on success rate. The dashed line illustrates the mean success rate of converged PVDR. The solid line and the shaded regions represent the mean and variance of performance across five runs with different seeds. \textit{PVDR w/o prior opt}, \textit{PVDR w/o act sl}, \textit{PVDR w/o act rl}, and \textit{PVDR random} refer to PVDR without goal-oriented term in $\mathcal{L}_{prior}$, PVDR without $\mathcal{L}_{act}$, PVDR without PPO loss, and PVDR with random representation.}
    \label{fig:mw.ablation}
\end{figure}

\noindent\textbf{Ablations.} 
We verify the effectiveness of PVDR’s design through ablation studies. First, we replace the selected visual dynamics representation with a random tensor to verify whether the agent utilizes the effective information in visual dynamics representations for decision-making. In addition, considering multiple loss terms are used during the online adaptation, we design ablation experiments on four tasks in Meta-World to verify the effectiveness of these loss terms. Specifically, we conduct ablation studies separately for $\mathcal{L}_{prior}$, $\mathcal{L}_{act}$, and the PPO loss used to optimize the alignment of actions. The learning curves are shown in Figure~\ref{fig:mw.ablation}. The experimental results indicate that visual dynamics representations indeed capture conducive information, and all these optimization terms contribute positively to the PVDR's performance. 
More ablations and additional experiments are conducted in Appendix~\ref{app:Add_Exp}.

\begin{figure}[tb]
    \centering
    \includegraphics[width=0.75\linewidth]{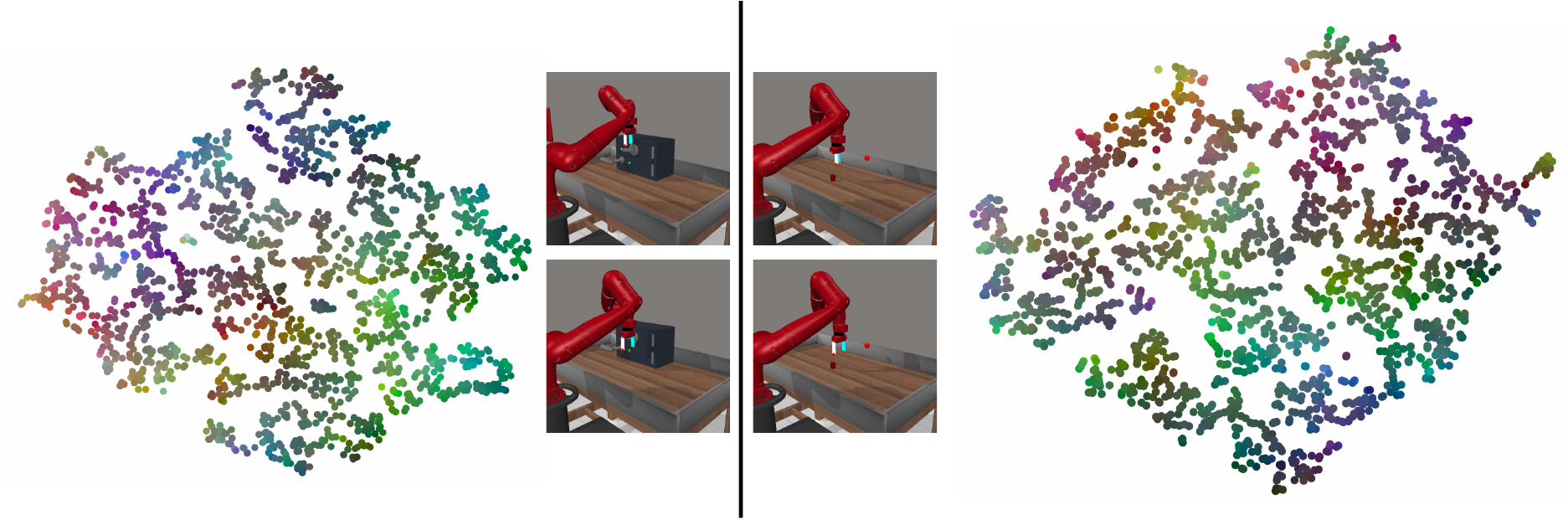}
    \caption{Two visualization cases of relevance between visual dynamics representations and corresponding actions under certain observation contexts. 2000 visual dynamics representations are sampled from the prior in each case and mapped to a 2-dimensional space via t-SNE. Each 4-dimensional action vector is scaled from $[-1,1]$ box to $[0,1]$ box and the scaled 4-dimensional vector is viewed as the CMYK color of the corresponding visual dynamics representation.}
    \label{fig:mw.visualization}
\end{figure}

\noindent\textbf{Visualization.}
In addition to assessing the performance on downstream tasks, we also seek to showcase whether the visual dynamics representations could 
be effectively aligned with actions. We examine the correlation between representations and actions through visualization. Specifically, under a certain observation context $\mathbf{x}_t$ in the task, we sample 2000 visual dynamics representations $\{z^i_t \sim P_\theta(\cdot \mid \mathbf{x}_t) \}_{i=1}^{2000}$ from the prior $P_\theta$. The $z^i_t$ are then labeled with the corresponding actions through the action alignment module $\Pi_\phi$, constructing the set $\{(z^i_t, a^i_t \sim \Pi_\phi(\mathbf{x}_t, z^i_t))\}_{i=1}^{2000}$. Based on this set, $z^i_t$ is mapped to a point in the 2-dimensional space via t-SNE, and each point is colored according to the value of the action $a^i_t$. The visualization results in two cases are shown in Figure~\ref{fig:mw.visualization}. The continuity in the distribution of colors can be observed on the two-dimensional plane, suggesting a correlation between visual dynamics representations and actions in these observation contexts. 

\begin{figure}[tb]
    \centering
    \includegraphics[width=0.75\linewidth]{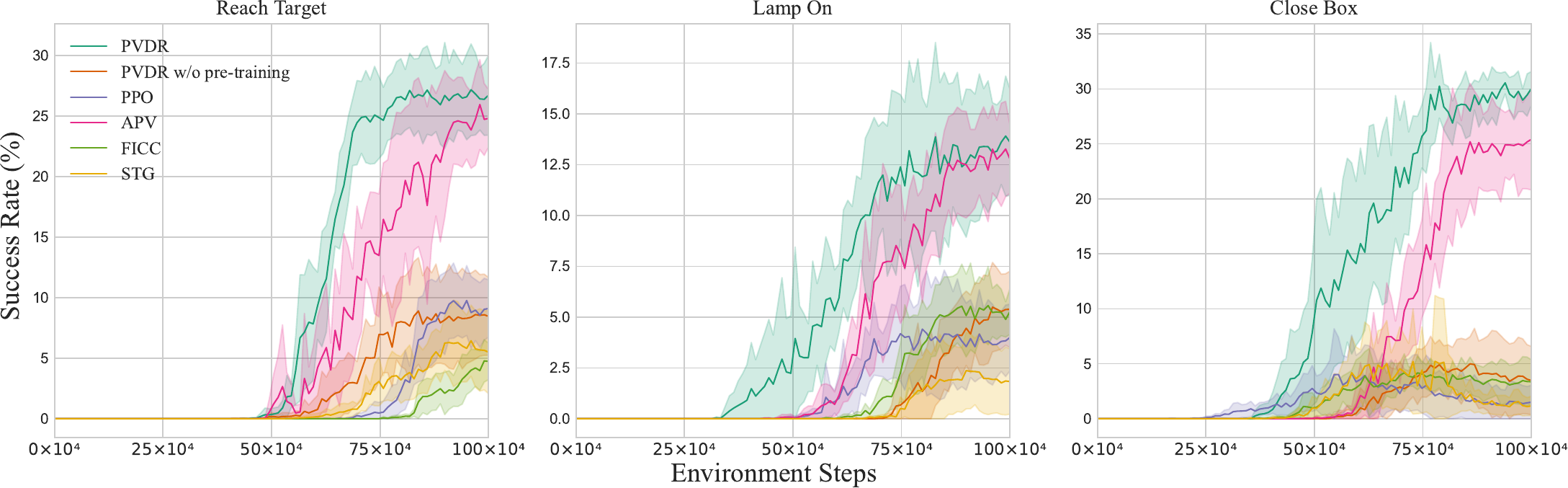}
    \caption{Learning curves of PVDR compared with five baselines on three RLBench tasks measured on success rate. The solid line and shaded regions represent the mean and variance of the performance across five runs with different seeds.}
    \label{fig:RLB.baseline}
\end{figure}

\subsection{RLBench Experiments}
We also examine the effectiveness of PVDR in more varying and challenging tasks from the RLBench environment. We run PVDR and baselines on three tasks from RLBench and the learning curves of success rates are shown in Figure~\ref{fig:RLB.baseline}. PVDR also outperforms the baselines in these three tasks, and the experimental results in RLBench further validate the effectiveness of PVDR.

%% file: content/06_conclusion.tex
\section{Conclusion}
\label{sec:conclusion}
In this paper, we propose PVDR to address the challenge of \textit{pre-training with videos}. PVDR bridges the gap between pre-training videos and downstream tasks with visual dynamics representations. PVDR contains a visual dynamics model and an action alignment module. As a two-stage visual control algorithm, PVDR pre-trains visual dynamics representations with the video prediction task and adapts the visual dynamics model to the downstream tasks. The representations can be aligned with actions to enable executable policy via online learning. The experimental results in a series of visual control tasks demonstrate that the visual dynamics representations are appropriate for pre-training with domain-gap videos, and the design for online adaptation and alignment has proved effective. In addition, a discussion of the limitation of PVDR can be found in 
Appendix~\ref{app:limit}.


%% file: content/App_Implementation.tex
\section{Experiments Details}
\label{app:experiments}
In this section, we describe the details of the implementations (\Cref{app:imple}), the hyperparameters (\Cref{app:hyper}), and the task settings (\Cref{app:task}).
\subsection{Implementations}
\label{app:imple}

We build our framework based on PyTorch~\cite{paszke2019pytorch} and use the implementations of Transformer tricks from the codebase x-transformers\footnote{https://github.com/lucidrains/x-transformers}. We pre-train a VQGAN with the images extracted from BAIR and downstream tasks based on the released code\footnote{https://github.com/CompVis/taming-transformers}. We use  8$\times$A100 Nvidia GPU and 64 CPU cores for the pre-training run, while using one single A100 Nvidia GPU and 16 CPU cores for each training run. 10 hours are required to pre-train PVDR. For the online stage, 12 hours are required to train on each Meta-World~\cite{yu2020meta} task, and 18 hours are required to train on each RLBench~\cite{james2020rlbench} task. The details of the implementations are as follows.

\begin{figure}[tb]
    \centering
    \setlength{\abovecaptionskip}{2mm}
    \includegraphics[width=0.85\linewidth]{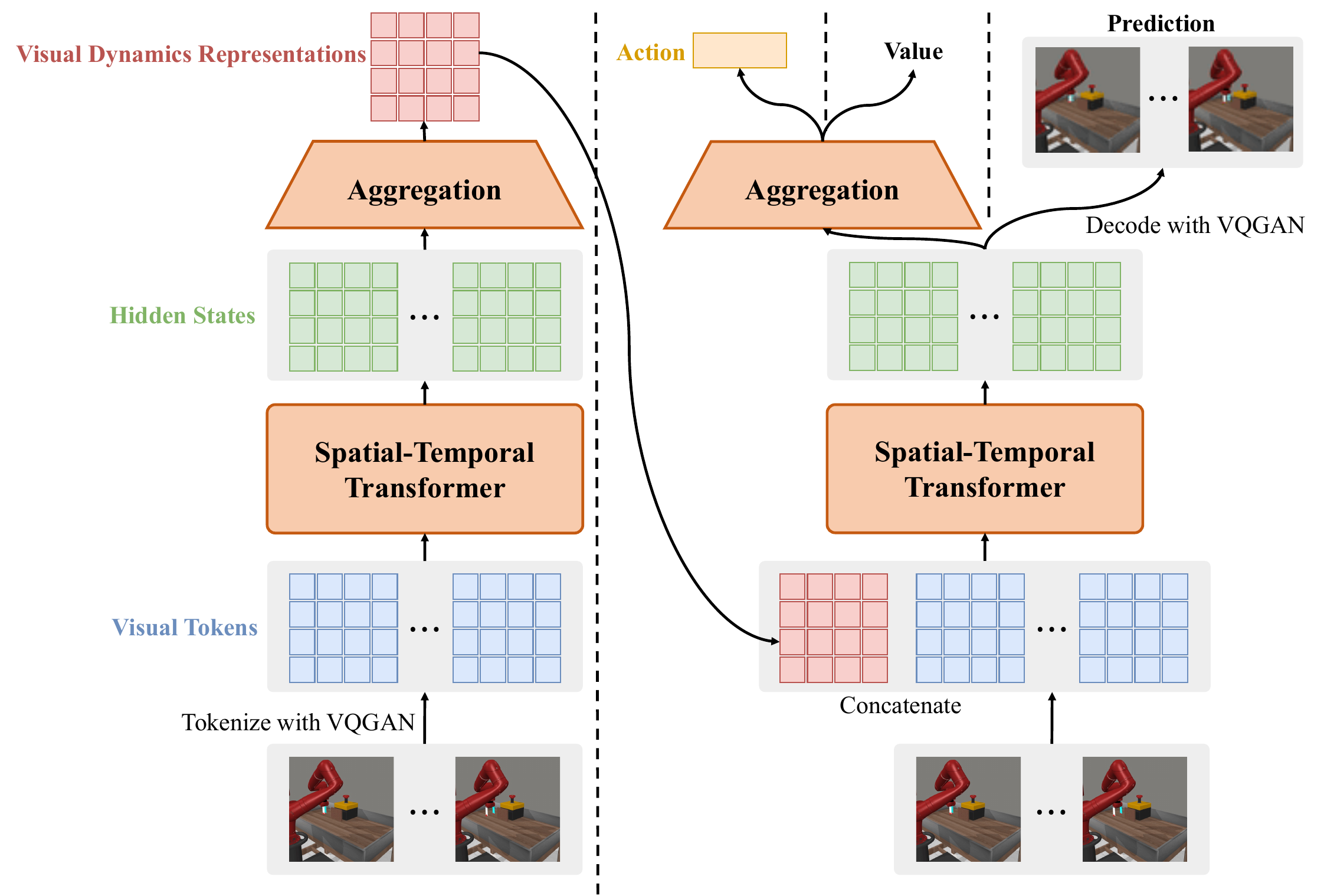}
    \caption{Illustration of module structures in PVDR. The visual dynamics encoder and the visual dynamics prior share the structure on the left. The visual dynamics decoder, the action alignment module, and the critic network share the main network on the right, with distinct post-processing forms of the hidden states.}
    \label{fig:structure}
\end{figure}

\paragraph{PVDR Structure.} We implement all modules in PVDR as Transformer-based structures. Overall, the structures of the modules in PVDR, including a visual dynamics encoder, a visual dynamics prior, a visual dynamics decoder, an action alignment module, and a critic network, are shown in \Cref{fig:structure}. These modules use a pre-trained VQGAN to compress the raw frames into grids of visual tokens and subsequently employ the Spatial-Temporal Transformers~\cite{xu2020spatial} to process spatial-temporal visual token sequences. The modules differ from each other in the input data type and the post-processing form of the hidden states. Specifically, the visual dynamics representations are in the same shape as visual token grids and are viewed as grids of prompts that will be concatenated with the visual token grids. During the decoding process, the hidden state of each visual token will be projected onto a vector, whose dimension equals the size of the VQGAN codebook. The softmax value of each dimension is viewed as the probability of the corresponding code. For a quick decoding process, we do not use the beam search. Instead, we recurrently use the code with the highest probability to form a grid and directly decode one visual frame with VQGAN.

\begin{figure}[tb]
    \centering
    \setlength{\abovecaptionskip}{2mm}
    \includegraphics[width=0.7\linewidth]{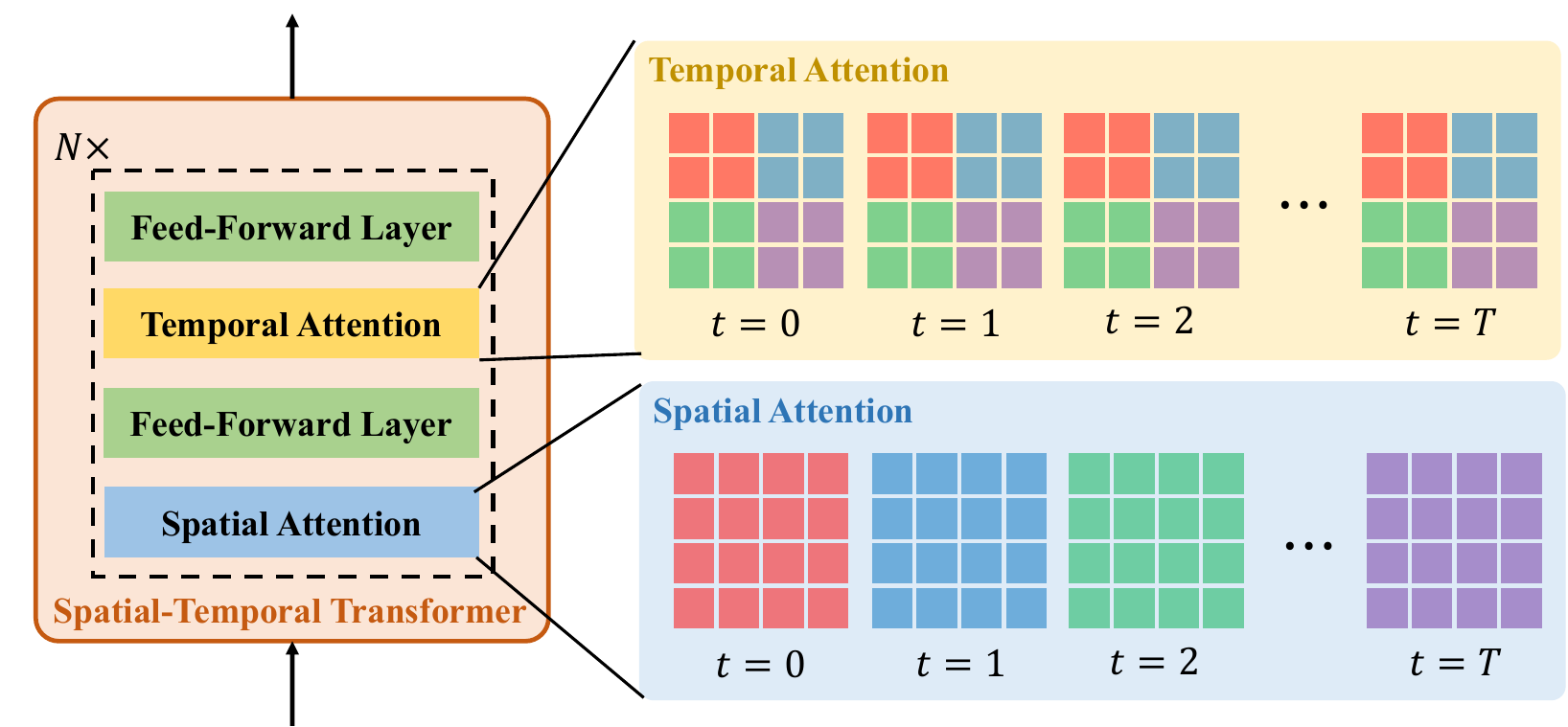}
    \caption{Illustration of Spatial-Temporal Transformer in PVDR. An ST Transformer (left) is composed of $N$ cross-stacked Spatial Attention and Temporal Attention blocks, with interleaved Feed-Forward Layers. The visual tokens (right) in the same color are processed in one attention map.}
    \label{fig:st_attn}
\end{figure}

\paragraph{ST Transformer.} In our PVDR implementation, the core structure is the Spatial-Temporal Transformer (ST Transformer). For a lower attention computation burden, an ST Transformer uses cross-stacked spatial and temporal attention blocks to, respectively, process the information in spatial and temporal sequences. Generally, we design spatial and temporal attention blocks following MaskViT~\cite{gupta2022maskvit}. As shown in \Cref{fig:st_attn}, the spatial attention blocks process the attention map of visual tokens at the same timestep, while the temporal attention blocks process the attention map of visual tokens in a small local spatial window alongside the temporal sequence. 

\noindent Formally, $(l+m)\times h\times w$ visual tokens are processed by the ST Transformer, which are numbered $\{\mathcal{E}_{i,j,k}\mid 0\le i <h, 0\le j<w, 0\le k < l+m \}$. Take a certain visual token, $\mathcal{E}_{i^\star,j^\star,k^\star}$, for example. The visual tokens in the set $\{\mathcal{E}_{i,j,k^\star}\mid0\le i <h, 0\le j<w\}$ are included in its spatial attention map. And the visual tokens in the set $\{\mathcal{E}_{i,j,k} \mid  \lfloor\frac{i}{b}\rfloor=\lfloor\frac{i^\star}{b}\rfloor, \lfloor\frac{j}{d}\rfloor=\lfloor\frac{j^\star}{d}\rfloor, 0\le k < l+m \}$ are included in the temporal attention map, where $b\times d$ are the local attention window size. Additionally, the temporal attention block used in the visual dynamics decoder employs a causal mask. That is, the visual tokens in the set $\{\mathcal{E}_{i,j,k} \mid \lfloor\frac{i}{b}\rfloor=\lfloor\frac{i^\star}{b}\rfloor, \lfloor\frac{j}{d}\rfloor=\lfloor\frac{j^\star}{d}\rfloor, 0\le k < k^\star \}$ are included in the causal-masked temporal attention calculation of $\mathcal{E}_{i^\star,j^\star,k^\star}$.

\begin{table}[t!]
    \begin{center}
    \caption{Hyperparameters of the VQGAN used in PVDR.}
    \vspace{-2mm}
    \label{tab:vqgan}
    \setlength{\tabcolsep}{4pt}
    \begin{tabular}{lc}
    \toprule
        Parameter & Setting  \\
    \midrule
        Resolution & $128\times128$ \\
        Visual token grid size & $16\times16$ \\
    \midrule
        Codebook size & $256$ \\
        Code dimension & $1024$ \\
        Codebook loss weight & $1.0$ \\
    \midrule
        Discriminator loss weight & $0.8$ \\
        Discriminator loss start step & $10000$ \\
    \midrule
        Number of downsampling blcoks & $4$ \\
        Number of residual blocks & $2$ \\
        Channel multiplier & $(1,2,2,4)$ \\
        Channel in \& out & in:$160$ out:$64$\\
    \midrule
        Minibatch size & $1600$ \\
        Optimizer & Adam \\
        Optimizer: learning rate & $4.5\text{e}-6$\\
        Dropout rate & $0.1$ \\
        Training steps & $4\text{e}5$ \\
    \bottomrule
    \end{tabular}
    \end{center}
\vspace{-4mm}
\end{table}

\paragraph{Baselines.} For PVDR without pre-training and PVDR-based algorithms for ablation, the structures of the modules are consistent as described above. For PPO, the actor network is in the same structure as the action alignment module in PVDR, while the critic network is identical to the one in PVDR. The setups of APV, FICC, and STG are consistent with the original setups. In particular, FICC requires a discrete action space for search. Thus, we evenly divide the continuous action space of environments into 16 subspaces and uniformly sample one action in each space for each round of search. As PVDR does not learn a reward function, we replace the reward function in FICC with $r_t = -\left\|o_{t+1} - g\right\|_1$ during inference for a fair comparison. All algorithms use $r_t = -\left\|o_{t+1} - g\right\|_1$ as an intrinsic reward for the goal-conditioned setting.

\subsection{Hyperparameters}
\label{app:hyper}
In this section, we list the hyperparameters in our PVDR implementation. The hyperparameters of the pre-trained VQGAN we used are shown in \Cref{tab:vqgan}. The hyperparameters on structures and learning of ST Transformers in PVDR are shown in \Cref{tab:trans}. The hyperparameters of the PPO training are shown in \Cref{tab:ppo}. In fact, the hyperparameters of the PPO baseline are identical to those shown in \Cref{tab:ppo}. In particular, there are slight differences in our PPO training of different Meta-World tasks. Here, we list the tasks of 3 different hyperparameter sets. \textbf{Task Set 1}: button press topdown, dial turn, drawer close, reach, window open. \textbf{Task Set 2}: button press, plate slide, plate slide back,  window close. \textbf{Task Set 3}: plate slide side, hand insert, drawer open.

\begin{table}[t!]
    \begin{center}
    \caption{Hyperparameters of the ST Transformer used in PVDR.}
    \vspace{-2mm}
    \label{tab:trans}
    \setlength{\tabcolsep}{4pt}
    \begin{tabular}{lc}
    \toprule
        Parameter & Setting  \\
    \midrule
        Visual token grid size & $16\times16$\\
        Local window size & $4\times 4$ \\
        Context frames length $m$ & $2$ \\
        Future frames length $l$ & $6$ \\
    \midrule
        Visual dynamics representation shape & $16\times16\times32$ \\
        Attention dimension & $512$ \\
        Feedforward dimension & $512$ \\
        Token embedding dimension & $1024$ \\
        Visual token vocabulary size & $256$ \\
    \midrule
        Encoder Transformer depth & $6$ \\
        Encoder Transformer heads & $4$\\
        Prior Transformer depth & $6$\\
        Prior Transformer heads & $4$\\
        Decoder Transformer depth & $3$\\
        Decoder Transformer heads & $4$\\
        Action alignment Transformer depth & $3$\\
        Action alignment Transformer heads & $4$\\
        Critic Transformer depth & $3$\\
        Critic Transformer heads & $4$\\
    \midrule
        Loss weight $\lambda_1$ & $1\text{e}-3$\\
        Loss weight $\lambda_2$ & $2.5$ (RLBench); $0.8$ (Meta-World)\\
        Reward weight $\lambda_3$ & $4.5$ (RLBench); $1.0$ (Meta-World)\\
    \midrule
        Minibatch size & $1025$ \\
        Optimizer & Adam \\
        Pre-training learning rate & $1\text{e}-4$\\
        Fine-tuning learning rate & $1\text{e}-5$\\
        Downstream training rate & $4\text{e}-5$ \\
        Pre-training steps & $3.5\text{e}5$ \\
    \bottomrule
    \end{tabular}
    \end{center}
\vspace{-4mm}
\end{table}

\begin{table}[t!]
    \begin{center}
    \caption{Hyperparameters of PPO in PVDR.}
    \vspace{-2mm}
    \label{tab:ppo}
    \setlength{\tabcolsep}{4pt}
    \begin{tabular}{lcccc}
    \toprule
        Parameter & RLBench & Task Set 1 & Task Set 2 & Task Set 3 \\
    \midrule
        $\lambda$ & \multicolumn{4}{c}{$0.92$}\\
        $\gamma$ & \multicolumn{4}{c}{$0.99$}\\
        $\epsilon$ & $0.4$ & $0.6$ & $0.2$ & $0.5$\\
        $\epsilon_{value}$ & \multicolumn{4}{c}{$10$}\\
        $c_{entropy}$ & $3\text{e}-4$ & $1\text{e}-2$ & $2\text{e}-4$ & $5\text{e}-3$\\
        $c_{value}$ & \multicolumn{4}{c}{$0.5$}\\
        Max gradient norm &  \multicolumn{4}{c}{$1.0$}\\
        Minibatch size & \multicolumn{4}{c}{$100$}\\
        Actor learning rate &  \multicolumn{4}{c}{$3\text{e}-4$}\\
        Critic learning rate &  \multicolumn{4}{c}{$1\text{e}-3$}\\
        Training Epochs & $500$ & $200$ & $200$ & $200$\\
    \bottomrule
    \end{tabular}
    \end{center}
    \vspace{-4mm}
\end{table}

\subsection{Task Settings}
\label{app:task}
\begin{figure}[tb]
    \centering
    \setlength{\abovecaptionskip}{2mm}
    \includegraphics[width=0.7\linewidth]{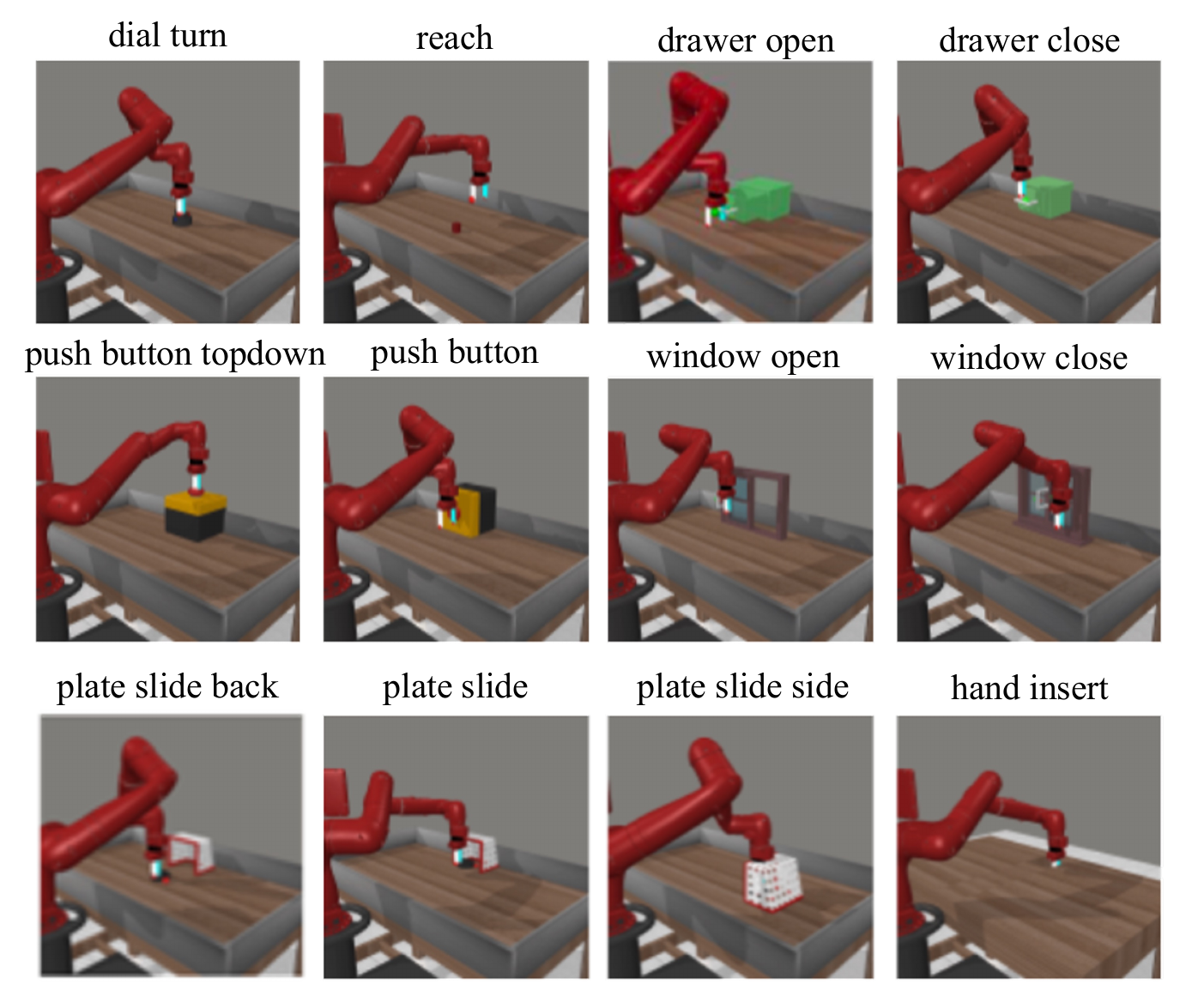}
    \caption{Illustration of the image goals in Meta-World tasks.}
    \label{fig:meta.goal}
\end{figure}

\begin{figure}[b!]
    \centering
    \setlength{\abovecaptionskip}{2mm}
        \includegraphics[width=0.5\linewidth]{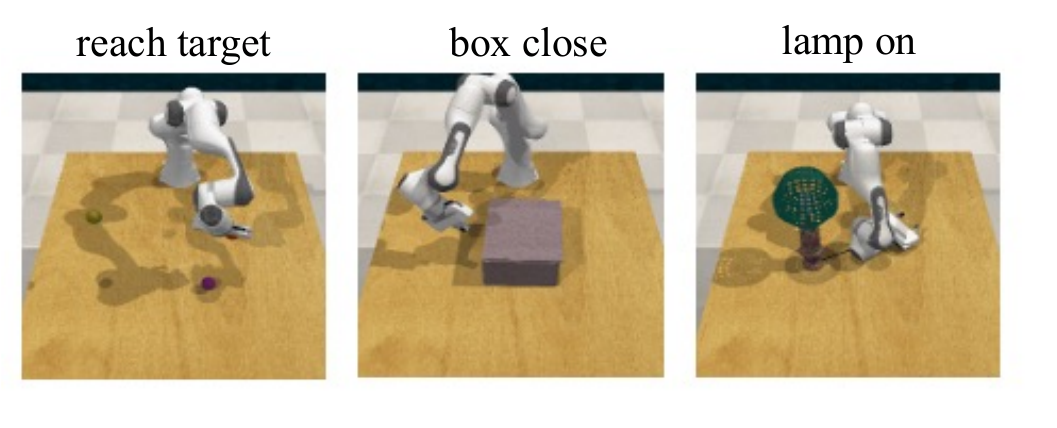}
    \caption{Illustration of the image goals in RLBench tasks.}
    \label{fig:rlbench.goal}
\end{figure}

We select 12 tasks from Meta-World and 3 tasks from RLBench for our experiments. As we consider goal-based tasks, we replace the original dense reward in Meta-World with the sparse reward. A positive $r_{suc}$ will be given when the goal is achieved. Concretely, $r_{suc}$ is $100$ in RLBench tasks and $60$ in Meta-World tasks.
And the visual goals of the tasks are shown in \Cref{fig:meta.goal,fig:rlbench.goal}.

\clearpage

%% file: content/App_Samples.tex
\newpage
\section{Additional Experiments}
\label{app:Add_Exp}
\subsection{Hyperparameters Influence}
\begin{figure}[!h]
    \setlength{\abovecaptionskip}{2mm}
    \setlength{\belowcaptionskip}{-6mm}
    \centering
    \includegraphics[width=0.9\linewidth]{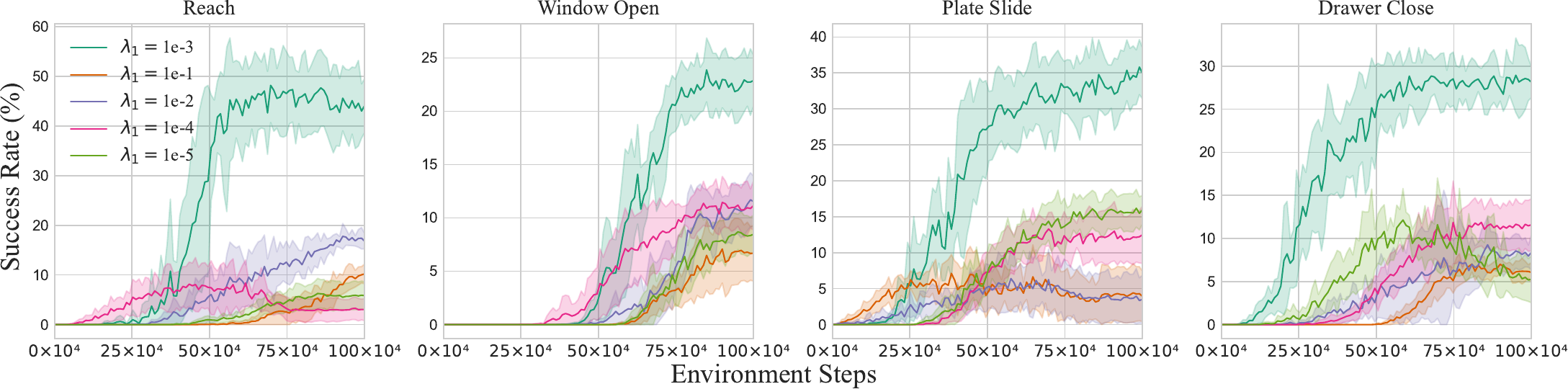}
    \caption{Learning curves of different $\lambda_1$ value in PVDR on four Meta-World tasks measured on success rate. The solid line and the shaded regions represent the mean and variance of performance across five runs with different seeds.}
    \label{fig:factor_1}
\end{figure}

\begin{figure}[!h]
    \setlength{\abovecaptionskip}{2mm}
    \setlength{\belowcaptionskip}{-6mm}
    \centering
    \includegraphics[width=0.9\linewidth]{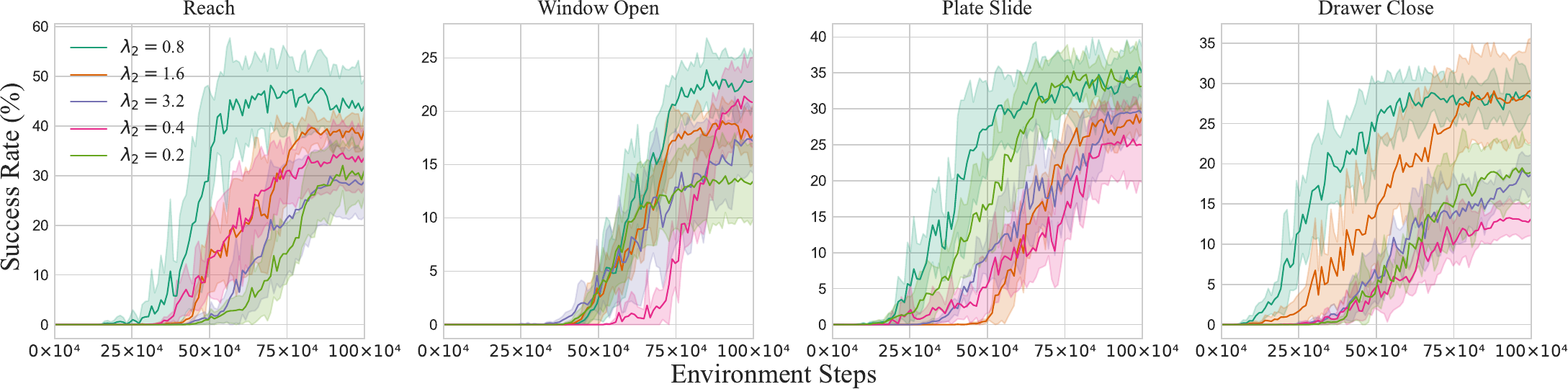}
    \caption{Learning curves of different $\lambda_2$ value in PVDR on four Meta-World tasks measured on success rate. The solid line and the shaded regions represent the mean and variance of performance across five runs with different seeds.}
    \label{fig:factor_2}
\end{figure}

\begin{figure}[!h]
    \setlength{\abovecaptionskip}{2mm}
    \setlength{\belowcaptionskip}{-6mm}
    \centering
    \includegraphics[width=0.9\linewidth]{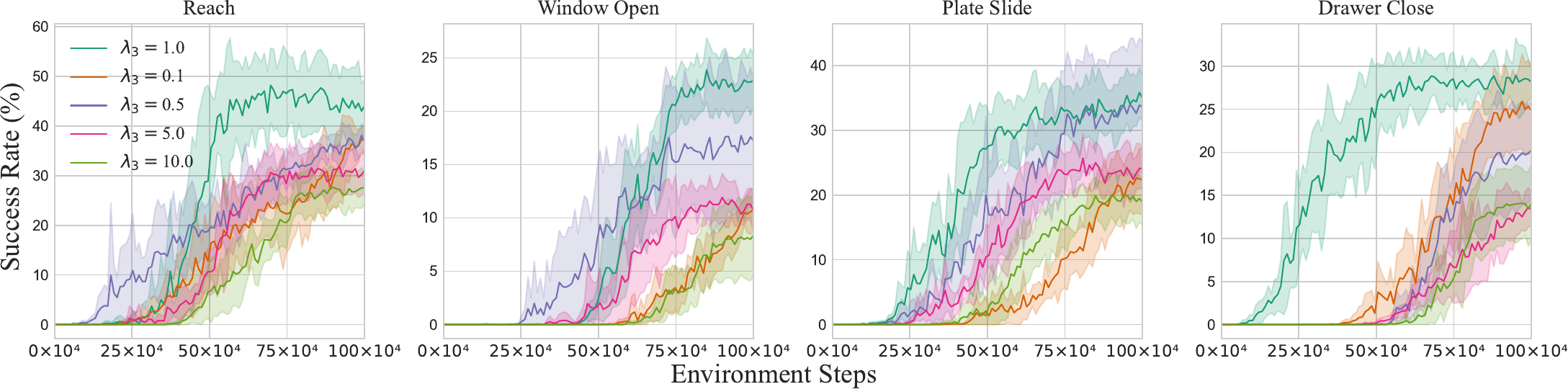}
    \caption{Learning curves of different $\lambda_3$ value in PVDR on four Meta-World tasks measured on success rate. The solid line and the shaded regions represent the mean and variance of performance across five runs with different seeds.}
    \label{fig:factor_3}
\end{figure}

Given that PVDR contains three weight factor hyperparameters ($\lambda_1,\lambda_2,\lambda_3$), we conduct additional experiments to explore the impact of these factors on the experimental results. Specifically, we experiment with a range of values for these three factors on four Meta-World tasks, and the learning curves are shown in \Cref{fig:factor_1,fig:factor_2,fig:factor_3}. The performance of PVDR is shown to be quite sensitive to the value of $\lambda_1$, relatively stable to the value of $\lambda_2$, and moderately influenced by the value of $\lambda_3$.

\clearpage

In addition, we conduct ablation studies on the horizon length ($m+n$) in PVDR. As shown in \Cref{tab:horizon}, we find that shorter horizons negatively impact the performance. This indicates that encoding more abstract and compressed information from longer horizons is beneficial.

\begin{table}[t]
    \centering
    \setlength{\tabcolsep}{4pt}
    \caption{Success rates ($\%$) of PVDR with various horizon lengths in 2 Meta-World tasks. The means and variances of the performance over five runs with different seeds are reported.}
    \begin{tabular}{c c c c}
    \toprule
    Horizon Length ($\mathbf{m+n}$) & $\mathbf{2+6}$   & $\mathbf{1+1}$ & $\mathbf{2+2}$\\ 
    \midrule
    Reach & $45.0 \pm 1.3$ & $7.8 \pm 1.5$ & $11.0 \pm 4.1$ \\ 
    Plate Slide & $35.2 \pm 4.2$ & $7.6 \pm 2.2$ & $11.4 \pm 3.4$\\ 
    \bottomrule
    \end{tabular}
    \label{tab:horizon}
\end{table}

\subsection{Various Dataset}

We also conduct additional experiments to evaluate the generalization ability of PVDR using various datasets: trajectory dataset collected in Meta-World, Open-X Embodiment datasets~\cite{open2024open} (USC Jaco Pay and Berkeley Cable Routing). The size of each used dataset is consistent with the BAIR dataset to ensure fairness. For the four datasets (including BAIR) we conduct experiments on, they represent different quality of the pre-training datasets:(1) \textbf{Meta-World Dataset} is in-domain. (2) \textbf{BAIR Dataset} is task-relevant but has a visual gap. (3) \textbf{USC Jaco Pay Dataset} uses Jaco robot in less relevant tasks. (4) \textbf{Berkeley Cable Routing Dataset} uses Franka robot in totally different tasks. The results are shown in

\Cref{tab:datasets}, indicating different pre-training datasets impact the performance and generalization, and PVDR works effectively across various datasets. The benefits of pre-training are limited by the increasing gap, which necessitates capturing the more generalizable dynamics priors. Our results indicate that PVDR effectively bridges the gap across different pre-training datasets and downstream tasks.

\begin{table}[t]
    \centering
    \setlength{\tabcolsep}{4pt}
    \caption{Success rates ($\%$) of PVDR in Meta-World with 4 various pre-training datasets. The datasets are represented with the following numbers: (1) \textbf{Meta-World Dataset}, (2) \textbf{BAIR Dataset}, (3) \textbf{USC Jaco Pay Dataset}, and (4) \textbf{Berkeley Cable Routing Dataset}. The means and variances of the performance over five runs with different seeds are reported.}
    \begin{tabular}{c c c c c c}
    \toprule
    \multicolumn{2}{c}{\textbf{Dataset}} & (1) & (2) & (3) & (4)\\
    \midrule
    \multirow{3}{*}{\textbf{Reach}} & PVDR & $\boldsymbol{65.4 \pm 5.9}$ & $\boldsymbol{45.0 \pm 1.3}$ & $\boldsymbol{47.2 \pm 3.2}$ & $\boldsymbol{39.0 \pm 4.1}$\\
    & APV & $64.2 \pm 4.4$ & $26.8 \pm 3.4$ & $35.8 \pm 2.6$ & $21.2 \pm 1.6$ \\
    & FICC &  $36.4 \pm 3.8$ & $5.4 \pm 2.3$ & $4.2 \pm 2.8$ & $0.0 \pm 0.0$ \\
    \midrule
    \multirow{3}{*}{\textbf{Plate Slide}} & PVDR & $\boldsymbol{56.0 \pm 2.5}$ & $\boldsymbol{35.2 \pm 4.2}$ & $\boldsymbol{30.8 \pm 2.7}$ & $\boldsymbol{22.2 \pm 2.2}$ \\
    & APV & $38.8 \pm 1.6$ & $20.4 \pm 2.6 $& $26.6 \pm 2.2$ & $16.8 \pm 0.8$ \\
    & FICC & $22.4 \pm 3.7$ & $5.2 \pm 3.7$ & $3.0 \pm 1.1$ & $0.0 \pm 0.0$ \\
    \midrule
    \multirow{3}{*}{\textbf{Hand Insert}} & PVDR & $\boldsymbol{45.8 \pm 3.7}$ & $\boldsymbol{28.6 \pm 3.0}$ & $\boldsymbol{27.2 \pm 1.3}$ & $\boldsymbol{17.8 \pm 2.0}$ \\
    & APV & $43.6 \pm 2.4$ & $27.6 \pm 1.4$& $26.0 \pm 0.6$ & $12.6 \pm 6.1$ \\ 
    & FICC & $14.2 \pm 1.2$ & $4.6 \pm 2.0$ & $2.4 \pm 1.4$ & $0.0 \pm 0.0$ \\ 
    \bottomrule
    \end{tabular}
    \label{tab:datasets}
\end{table}

\subsection{Discussion about ContextWM}
In our baseline selection, we choose APV instead of ContextWM, which is somehow an extension of APV. ContextWM introduces a representation of the visual context to `facilitate knowledge transfer between distinct scenes' and thus enables the leverage of videos from multiple sources. This mechanism can be incorporated in PVDR by adding an extra condition (context representation) during the video prediction. We don't compare PVDR with ContextWM because BAIR dataset doesn't contain distinct scenes. Instead, the single-scene dataset may hinder the learning of meaningful context representation. We conduct additional experiments with ContextWM and PVDR with context representation. The results in \Cref{tab:context} show that incorporating context representation indeed worsens performance in the current setting.

\begin{table}[t]
    \vspace{6mm}
    \centering
    \setlength{\tabcolsep}{4pt}
    \caption{Success rates ($\%$) of PVDR, PVDR with context representation and ContextWM in 2 Meta-World tasks. The means and variances of the performance over five runs with different seeds are reported.}
    \label{tab:context}
    \begin{tabular}{c c c c}
    \toprule
    & PVDR  & PVDR with context representation & ContextWM \\ 
    \midrule
    Reach & $45.0 \pm 1.3$ & $31.0 \pm 2.4$ & $26.0 \pm 1.7$ \\ 
    Plate Slide & $35.2 \pm 4.2$ & $22.0 \pm 2.6$ & $20.6 \pm 2.2$ \\ 
    \bottomrule
\end{tabular}
\end{table}

\vspace{12mm}
\section{Cases Visualization}
\label{app:cases}
\begin{figure}[h!]
    \centering
    \setlength{\abovecaptionskip}{2mm}
    \setlength{\belowcaptionskip}{-6mm}
    \includegraphics[width=0.78\linewidth]{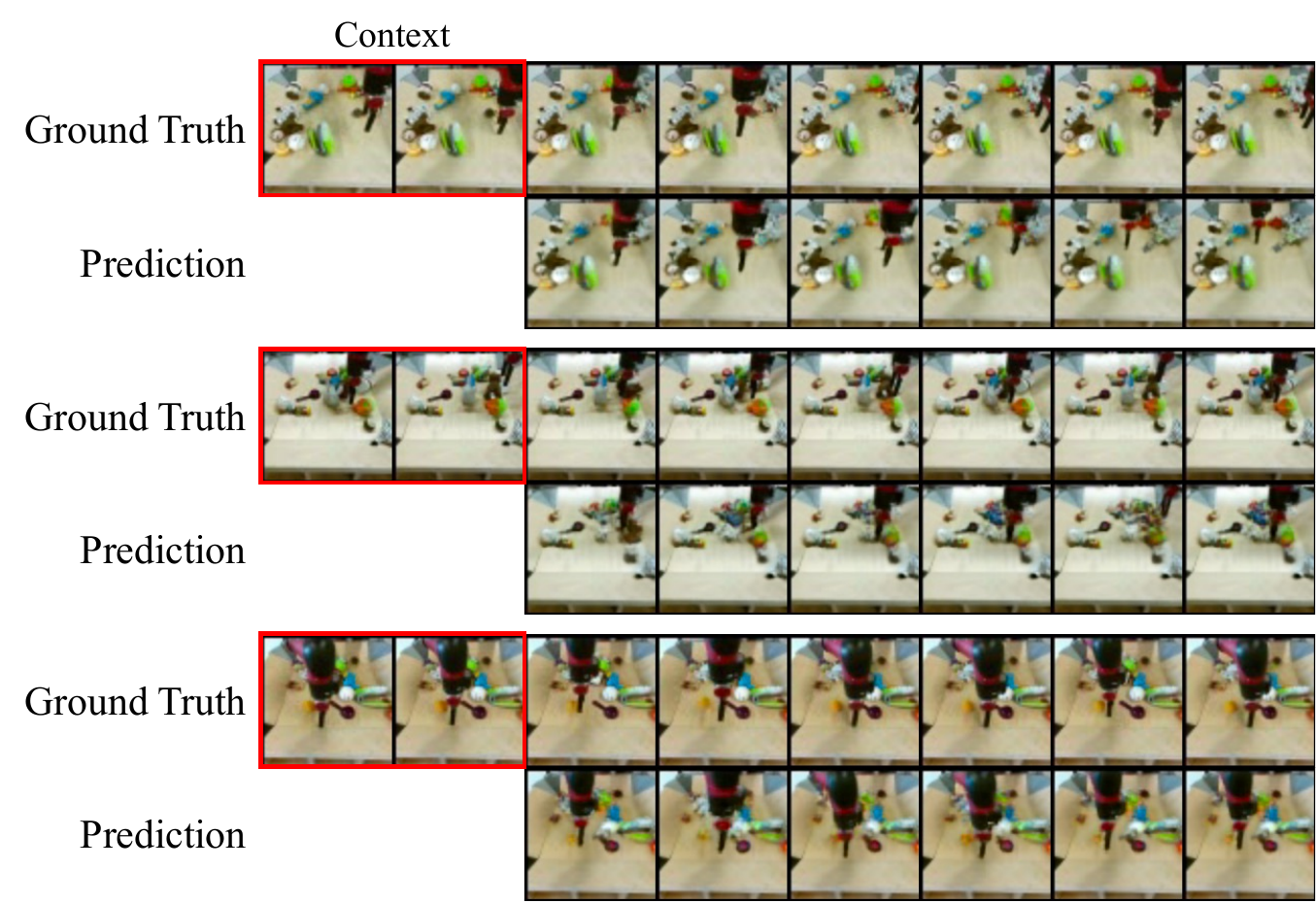}
    \caption{Future frames predicted by the pre-trained visual dynamics model on three cases from the BAIR dataset. The pre-trained model is capable of capturing the dynamics prior in the pre-trainig video datasets.}
    \label{fig:bair_case}
\end{figure}

\begin{figure}[h!]
    \centering
    \setlength{\abovecaptionskip}{2mm}
    \setlength{\belowcaptionskip}{-6mm}
    \includegraphics[width=0.85\linewidth]{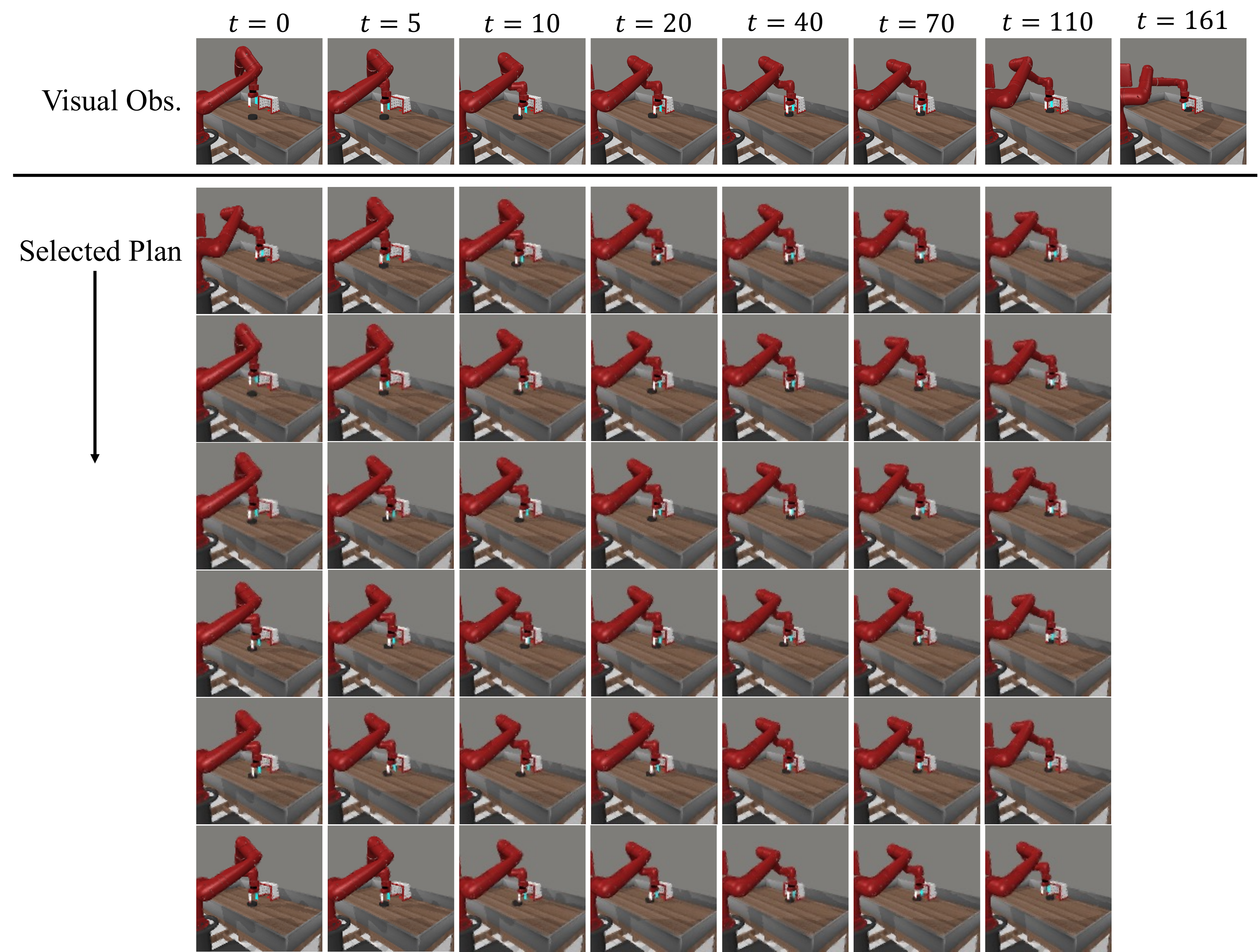}
    \caption{Visual observations and selected plans alongside the interaction of PVDR in the \textit{plate slide} task from Meta-World environment. PVDR is capable of generating meaningful plans with dynamics information and executing relevant actions.}
    \label{fig:meta_case}
\end{figure}

\begin{figure}[h!]
    \centering
    \setlength{\abovecaptionskip}{2mm}
    \setlength{\belowcaptionskip}{-6mm}
    \includegraphics[width=0.85\linewidth]{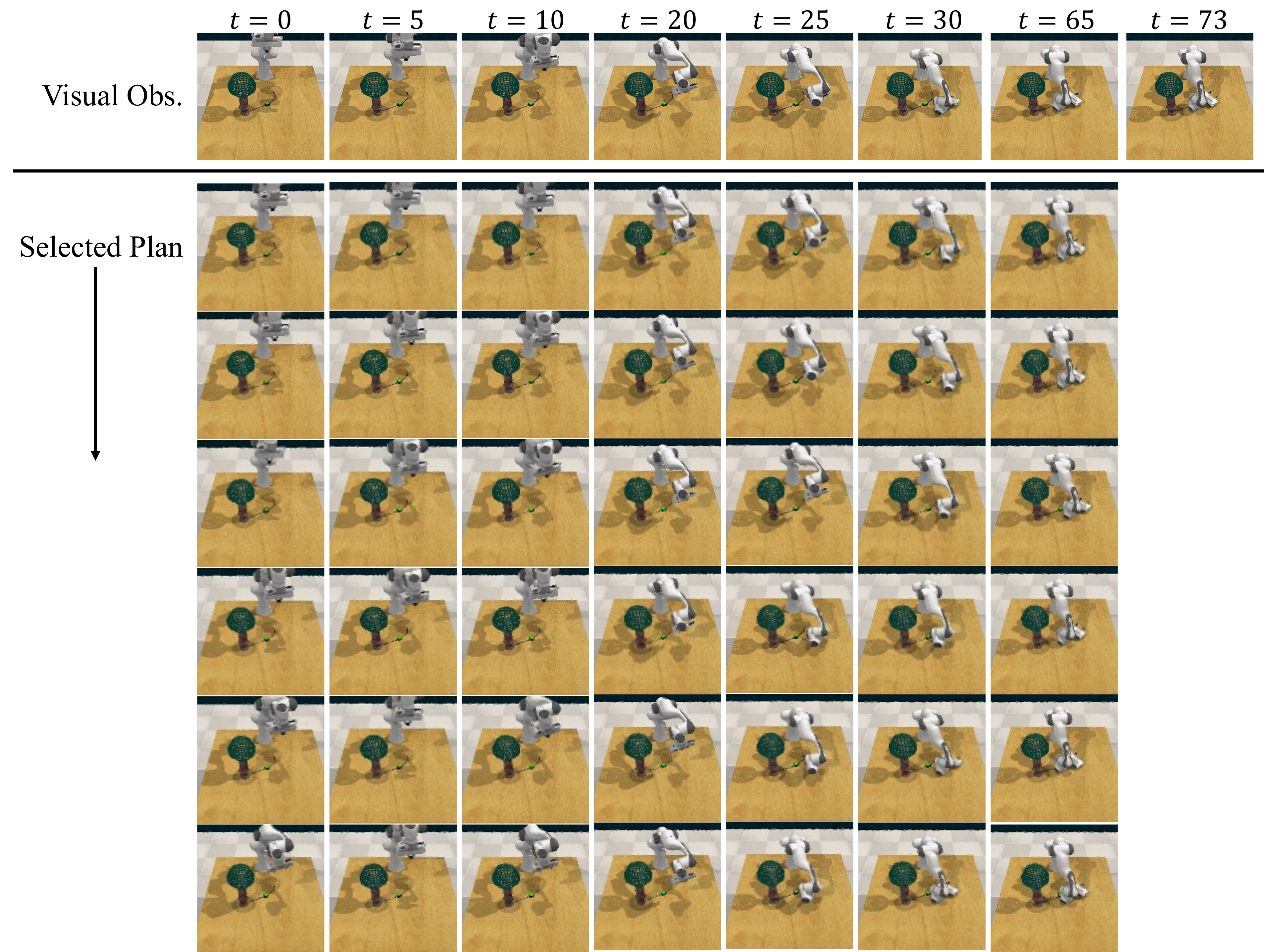}
    \caption{Visual observations and selected plans alongside the interaction of PVDR in the \textit{lamp on} task from RLBench environment. PVDR is capable of generating meaningful plans with dynamics information and executing relevant actions.}
    \label{fig:rlb_case}
\end{figure}

\begin{figure}[h!]
    \centering
    \setlength{\abovecaptionskip}{2mm}
    \setlength{\belowcaptionskip}{-6mm}
    \includegraphics[width=0.85\linewidth]{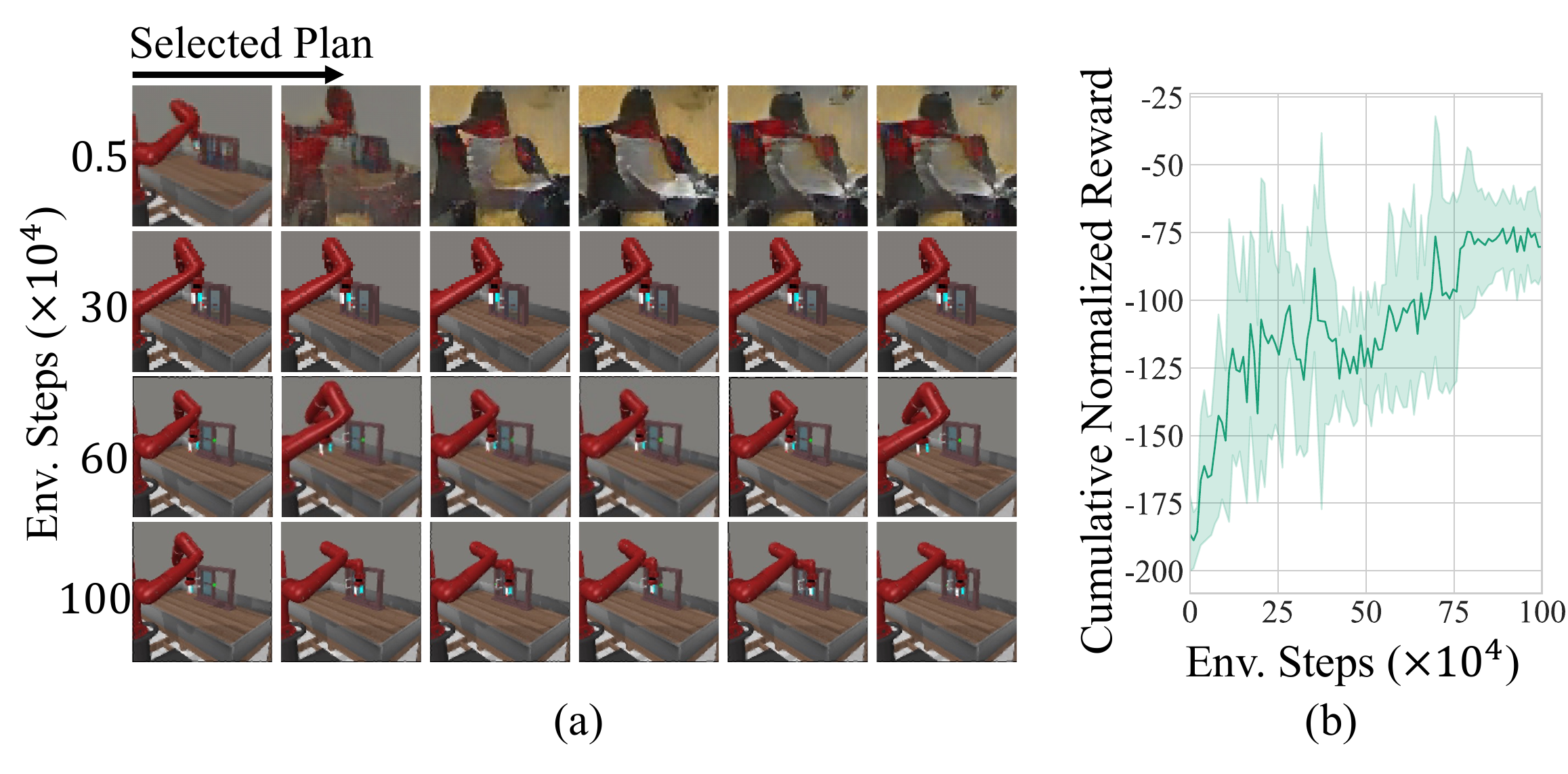}
    \caption{Examples from the adaptation in \textit{window close} task: selected visual plans (a) and action alignment reward curves (b).}
    \label{fig:online}
\end{figure}

In this section, we showcase the visualization of some cases from the pre-training video dataset and downstream tasks in \Cref{fig:bair_case,fig:meta_case,fig:rlb_case}. Additionally, we provide some examples in \Cref{fig:online}, showing the effectiveness of the online adaptation. As is shown in \Cref{fig:online}.(a), online adaptation helps to generate more reasonable visual plans. Furthermore, the gradually converging action alignment reward curves in \Cref{fig:online}.(b) indicate the effectiveness of the action alignment mechanism.
\clearpage

\clearpage

%% file: content/App_Extended.tex
\section{Extended Related Works}
Our work involves video prediction as well as goal-conditioned RL in addition to RL pre-training with videos. We are here to discuss the works related to these two topics. In addition, some works~\cite{chen2021learning,shao2021concept2robot,karamcheti2023language,wu2023unleashing,he2024large} pre-train modules with aligned video-text pairs to assist in solving tasks conditioned on textual goals, which is not within the scope of our work.

\label{app:extended}
\paragraph{Video Prediction.} As a complex amalgamation of visual comprehension and temporal sequence prediction, video prediction tasks aim to predict future frames given context frames. Generally, a series of generative models, prominently generative adversarial networks~\cite{clark2019adversarial,luc2020transformation}, variational autoencoders~\cite{villegas2019high,akan2022stochastic,walker2021predicting}, autoregressive models~\cite{yan2021videogpt,yan2023temporally,yu2023magvit}, and diffusion models~\cite{ho2022video,mei2023vidm,blattmann2023align}, have been shown instrumental in surmounting this challenge. Furthermore, many works~\cite{villegas2022phenaki,wang2024worlddreamer,singer2023makeavideo,blattmann2023align} consider the incorporation of text prompts/descriptions to aid in video generation. Our work mainly builds up the video prediction structure through a combination of earlier VAE-style works~\cite{villegas2019high,akan2022stochastic} and some recent works~\cite{xu2020spatial,gupta2022maskvit} based on spatial-temporal attention.

\paragraph{Goal-Conditioned Reinforcement Learning.} Our work uses the goal-conditioned RL framework. Unlike the setting based on a reward function, agents are provided with a behavior goal of the task in goal-conditioned RL. The goals can be images, feature vectors, language, commands, intrinsic skills, and so on, which 
are more general and flexible than the sometimes hard-to-define reward function. Many works address the challenge from various perspectives, such as optimization based on the divergence to the goal~\cite{schaul2015Uni,trott2019keeping,nair2020goal,mendonca2021discovering}, sub-goal generation/ selection~\cite{nair2018visual,Nair2020Hierarchical}, and relabeling techniques~\cite{andrychowicz2017hindsight}. Our work uses the most common and direct approach~\cite{ebert2017self,wu2018laplacian,pong2020skew} to measure the distance between observation and goal. Our choice accommodates the nature of the tasks in our experiments, and more delicate methods may help when handling more complex tasks.

\section{Limitation Discussion}
\label{app:limit}
Despite being an effective method for \textit{pre-training with videos}, PVDR has some limitations. Firstly, the learning of the reward function is not included in PVDR, and instead a preset reasonable reward function is assigned. Such a direct designation may not provide valid guidance for more complex tasks. In fact, some works have developed the usage of the foundation model as a reward function, which has the potential to be integrated to address the lack of a learned reward function. In addition, the pre-training videos are limited to a single source currently, and incorporating Internet videos from a wider range of sources may lead to a more general and capable agent. ContextWM~\cite{wu2023pre} actually already provides an effective way to extend single-source video data to large wide-source datasets through the introduction of context representations, and the introduction of this mechanism to PVDR may address this limitation.

%% file: camera-ready.bbl
\begin{thebibliography}{10}
\providecommand{\url}[1]{\texttt{#1}}
\providecommand{\urlprefix}{URL }
\providecommand{\doi}[1]{https://doi.org/#1}

\bibitem{akan2022stochastic}
Akan, A.K., Safadoust, S., G{\"u}ney, F.: Stochastic video prediction with structure and motion. arXiv preprint arXiv:2203.10528  (2022)

\bibitem{andrychowicz2017hindsight}
Andrychowicz, M., Wolski, F., Ray, A., Schneider, J., Fong, R., Welinder, P., McGrew, B., Tobin, J., Pieter~Abbeel, O., Zaremba, W.: Hindsight experience replay. In: Neural Information Processing Systems (2017)

\bibitem{aastrom1965optimal}
{\AA}str{\"o}m, K.J.: Optimal control of markov processes with incomplete state information i. Journal of Mathematical Analysis and Applications  \textbf{10},  174--205 (1965)

\bibitem{bahl2022human}
Bahl, S., Gupta, A., Pathak, D.: Human-to-robot imitation in the wild. arXiv preprint arXiv:2207.09450  (2022)

\bibitem{bahl2023affordances}
Bahl, S., Mendonca, R., Chen, L., Jain, U., Pathak, D.: Affordances from human videos as a versatile representation for robotics. In: Computer Vision and Pattern Recognition (2023)

\bibitem{baker2022video}
Baker, B., Akkaya, I., Zhokov, P., Huizinga, J., Tang, J., Ecoffet, A., Houghton, B., Sampedro, R., Clune, J.: Video pretraining (vpt): Learning to act by watching unlabeled online videos. Neural Information Processing Systems  (2022)

\bibitem{bhateja2023robotic}
Bhateja, C., Guo, D., Ghosh, D., Singh, A., Tomar, M., Vuong, Q., Chebotar, Y., Levine, S., Kumar, A.: Robotic offline rl from internet videos via value-function pre-training. arXiv preprint arXiv:2309.13041  (2023)

\bibitem{blattmann2023align}
Blattmann, A., Rombach, R., Ling, H., Dockhorn, T., Kim, S.W., Fidler, S., Kreis, K.: Align your latents: High-resolution video synthesis with latent diffusion models. In: Computer Vision and Pattern Recognition (2023)

\bibitem{bobrin2024align}
Bobrin, M., Buzun, N., Krylov, D., Dylov, D.V.: Align your intents: Offline imitation learning via optimal transport. arXiv preprint arXiv:2402.13037  (2024)

\bibitem{bojarski2016end}
Bojarski, M., Del~Testa, D., Dworakowski, D., Firner, B., Flepp, B., Goyal, P., Jackel, L.D., Monfort, M., Muller, U., Zhang, J., et~al.: End to end learning for self-driving cars. arXiv preprint arXiv:1604.07316  (2016)

\bibitem{brown2020language}
Brown, T., Mann, B., Ryder, N., Subbiah, M., Kaplan, J.D., Dhariwal, P., Neelakantan, A., Shyam, P., Sastry, G., Askell, A., et~al.: Language models are few-shot learners. Neural Information Processing Systems  (2020)

\bibitem{bruce2023learning}
Bruce, J., Anand, A., Mazoure, B., Fergus, R.: Learning about progress from experts. In: International Conference on Learning Representations (2023)

\bibitem{bruce2024genie}
Bruce, J., Dennis, M., Edwards, A., Parker-Holder, J., Shi, Y., Hughes, E., Lai, M., Mavalankar, A., Steigerwald, R., Apps, C., et~al.: Genie: Generative interactive environments. arXiv preprint arXiv:2402.15391  (2024)

\bibitem{chang2021learning}
Chang, M., Gupta, A., Gupta, S.: Learning value functions from undirected state-only experience. In: International Conference on Learning Representations (2022)

\bibitem{chen2021learning}
Chen, A.S., Nair, S., Finn, C.: Learning generalizable robotic reward functions from" in-the-wild" human videos. In: Robotics: Science and Systems (2022)

\bibitem{chen2020simple}
Chen, T., Kornblith, S., Norouzi, M., Hinton, G.: A simple framework for contrastive learning of visual representations. In: International Conference on Machine Learning (2020)

\bibitem{clark2019adversarial}
Clark, A., Donahue, J., Simonyan, K.: Adversarial video generation on complex datasets. arXiv preprint arXiv:1907.06571  (2019)

\bibitem{dosovitskiy2021an}
Dosovitskiy, A., Beyer, L., Kolesnikov, A., Weissenborn, D., Zhai, X., Unterthiner, T., Dehghani, M., Minderer, M., Heigold, G., Gelly, S., Uszkoreit, J., Houlsby, N.: An image is worth 16x16 words: Transformers for image recognition at scale. In: International Conference on Learning Representations (2021)

\bibitem{ebert2017self}
Ebert, F., Finn, C., Lee, A.X., Levine, S.: Self-supervised visual planning with temporal skip connections. Conference on Robot Learning  (2017)

\bibitem{edwards2019imitating}
Edwards, A., Sahni, H., Schroecker, Y., Isbell, C.: Imitating latent policies from observation. In: International Conference on Machine Learning (2019)

\bibitem{escontrela2024video}
Escontrela, A., Adeniji, A., Yan, W., Jain, A., Peng, X.B., Goldberg, K., Lee, Y., Hafner, D., Abbeel, P.: Video prediction models as rewards for reinforcement learning. Neural Information Processing Systems  (2023)

\bibitem{esser2021taming}
Esser, P., Rombach, R., Ommer, B.: Taming transformers for high-resolution image synthesis. In: Computer Vision and Pattern Recognition (2021)

\bibitem{ghosh2023reinforcement}
Ghosh, D., Bhateja, C.A., Levine, S.: Reinforcement learning from passive data via latent intentions. In: International Conference on Machine Learning (2023)

\bibitem{gupta2022maskvit}
Gupta, A., Tian, S., Zhang, Y., Wu, J., Mart{\'\i}n-Mart{\'\i}n, R., Fei-Fei, L.: Maskvit: Masked visual pre-training for video prediction. In: International Conference on Learning Representations (2022)

\bibitem{hafner2021mastering}
Hafner, D., Lillicrap, T.P., Norouzi, M., Ba, J.: Mastering atari with discrete world models. In: International Conference on Learning Representations (2021)

\bibitem{he2024large}
He, H., Bai, C., Pan, L., Zhang, W., Zhao, B., Li, X.: Large-scale actionless video pre-training via discrete diffusion for efficient policy learning. arXiv preprint arXiv:2402.14407  (2024)

\bibitem{he2022masked}
He, K., Chen, X., Xie, S., Li, Y., Doll{\'a}r, P., Girshick, R.: Masked autoencoders are scalable vision learners. In: Computer Vision and Pattern Recognition (2022)

\bibitem{ho2022video}
Ho, J., Salimans, T., Gritsenko, A., Chan, W., Norouzi, M., Fleet, D.J.: Video diffusion models. arXiv preprint arXiv:2204.03458  (2021)

\bibitem{james2020rlbench}
James, S., Ma, Z., Arrojo, D.R., Davison, A.J.: Rlbench: The robot learning benchmark \& learning environment. Robotics and Automation Letters  \textbf{5}(2),  3019--3026 (2020)

\bibitem{kaelbling1998planning}
Kaelbling, L.P., Littman, M.L., Cassandra, A.R.: Planning and acting in partially observable stochastic domains. Artificial Intelligence  \textbf{101}(1-2),  99--134 (1998)

\bibitem{karamcheti2023language}
Karamcheti, S., Nair, S., Chen, A.S., Kollar, T., Finn, C., Sadigh, D., Liang, P.: Language-driven representation learning for robotics. arXiv preprint arXiv:2302.12766  (2023)

\bibitem{kumar2020conservative}
Kumar, A., Zhou, A., Tucker, G., Levine, S.: Conservative q-learning for offline reinforcement learning. Neural Information Processing Systems  (2020)

\bibitem{li2024auxiliary}
Li, S., Han, S., Zhao, Y., Liang, B., Liu, P.: Auxiliary reward generation with transition distance representation learning. arXiv preprint arXiv:2402.07412  (2024)

\bibitem{luc2020transformation}
Luc, P., Clark, A., Dieleman, S., Casas, D.d.L., Doron, Y., Cassirer, A., Simonyan, K.: Transformation-based adversarial video prediction on large-scale data. arXiv preprint arXiv:2003.04035  (2020)

\bibitem{ma2022vip}
Ma, Y.J., Sodhani, S., Jayaraman, D., Bastani, O., Kumar, V., Zhang, A.: Vip: Towards universal visual reward and representation via value-implicit pre-training. arXiv preprint arXiv:2210.00030  (2022)

\bibitem{mei2023vidm}
Mei, K., Patel, V.: Vidm: Video implicit diffusion models. In: Association for the Advancement of Artificial Intelligence (2023)

\bibitem{menapace2021playable}
Menapace, W., Lathuiliere, S., Tulyakov, S., Siarohin, A., Ricci, E.: Playable video generation. In: Conference on Computer Vision and Pattern Recognition (2021)

\bibitem{mendonca2023structured}
Mendonca, R., Bahl, S., Pathak, D.: Structured world models from human videos. In: Robotics: Science and Systems (2023)

\bibitem{mendonca2021discovering}
Mendonca, R., Rybkin, O., Daniilidis, K., Hafner, D., Pathak, D.: Discovering and achieving goals via world models. In: Neural Information Processing Systems (2021)

\bibitem{nair2020contextual}
Nair, A., Bahl, S., Khazatsky, A., Pong, V., Berseth, G., Levine, S.: Contextual imagined goals for self-supervised robotic learning. In: Conference on Robot Learning (2020)

\bibitem{nair2018visual}
Nair, A.V., Pong, V., Dalal, M., Bahl, S., Lin, S., Levine, S.: Visual reinforcement learning with imagined goals. In: Neural Information Processing Systems (2018)

\bibitem{Nair2020Hierarchical}
Nair, S., Finn, C.: Hierarchical foresight: Self-supervised learning of long-horizon tasks via visual subgoal generation. In: International Conference on Learning Representations (2020)

\bibitem{nair2023r3m}
Nair, S., Rajeswaran, A., Kumar, V., Finn, C., Gupta, A.: R3m: A universal visual representation for robot manipulation. In: Conference on Robot Learning (2023)

\bibitem{nair2020goal}
Nair, S., Savarese, S., Finn, C.: Goal-aware prediction: Learning to model what matters. In: International Conference on Machine Learning (2020)

\bibitem{open2024open}
Padalkar, A., et\ al.: Open x-embodiment: Robotic learning datasets and rt-x models. arXiv preprint arXiv:2310.08864  (2023)

\bibitem{paszke2019pytorch}
Paszke, A., Gross, S., Massa, F., Lerer, A., Bradbury, J., Chanan, G., Killeen, T., Lin, Z., Gimelshein, N., Antiga, L., et~al.: Pytorch: An imperative style, high-performance deep learning library. In: Neural Information Processing Systems (2019)

\bibitem{pathak2018zero}
Pathak, D., Mahmoudieh, P., Luo, G., Agrawal, P., Chen, D., Shentu, Y., Shelhamer, E., Malik, J., Efros, A.A., Darrell, T.: Zero-shot visual imitation. In: Computer Vision and Pattern Recognition Workshops (2018)

\bibitem{peng2018sfv}
Peng, X.B., Kanazawa, A., Malik, J., Abbeel, P., Levine, S.: Sfv: Reinforcement learning of physical skills from videos. ACM Transactions On Graphics  \textbf{37}(6),  1--14 (2018)

\bibitem{pong2020skew}
Pong, V., Dalal, M., Lin, S., Nair, A., Bahl, S., Levine, S.: Skew-fit: State-covering self-supervised reinforcement learning. In: International Conference on Machine Learning (2020)

\bibitem{radford2021learning}
Radford, A., Kim, J.W., Hallacy, C., Ramesh, A., Goh, G., Agarwal, S., Sastry, G., Askell, A., Mishkin, P., Clark, J., et~al.: Learning transferable visual models from natural language supervision. In: International Conference on Machine Learning (2021)

\bibitem{radosavovic2023real}
Radosavovic, I., Xiao, T., James, S., Abbeel, P., Malik, J., Darrell, T.: Real-world robot learning with masked visual pre-training. In: Conference on Robot Learning (2023)

\bibitem{ranzato2014video}
Ranzato, M., Szlam, A., Bruna, J., Mathieu, M., Collobert, R., Chopra, S.: Video (language) modeling: a baseline for generative models of natural videos. arXiv preprint arXiv:1412.6604  (2014)

\bibitem{schaul2015Uni}
Schaul, T., Horgan, D., Gregor, K., Silver, D.: Universal value function approximators. In: International Conference on Machine Learning (2015)

\bibitem{schmeckpeper2021reinforcement}
Schmeckpeper, K., Rybkin, O., Daniilidis, K., Levine, S., Finn, C.: Reinforcement learning with videos: Combining offline observations with interaction. In: Conference on Robot Learning (2021)

\bibitem{schmidt2023learning}
Schmidt, D., Jiang, M.: Learning to act without actions. In: International Conference on Learning Representations (2024)

\bibitem{schulman2017proximal}
Schulman, J., Wolski, F., Dhariwal, P., Radford, A., Klimov, O.: Proximal policy optimization algorithms. arXiv preprint arXiv:1707.06347  (2017)

\bibitem{sekar2020planning}
Sekar, R., Rybkin, O., Daniilidis, K., Abbeel, P., Hafner, D., Pathak, D.: Planning to explore via self-supervised world models. In: International Conference on Machine Learning (2020)

\bibitem{seo2022reinforcement}
Seo, Y., Lee, K., James, S.L., Abbeel, P.: Reinforcement learning with action-free pre-training from videos. In: International Conference on Machine Learning (2022)

\bibitem{sermanet2018time}
Sermanet, P., Lynch, C., Chebotar, Y., Hsu, J., Jang, E., Schaal, S., Levine, S., Brain, G.: Time-contrastive networks: Self-supervised learning from video. In: International Conference on Robotics and Automation (2018)

\bibitem{shao2021concept2robot}
Shao, L., Migimatsu, T., Zhang, Q., Yang, K., Bohg, J.: Concept2robot: Learning manipulation concepts from instructions and human demonstrations. The International Journal of Robotics Research  \textbf{40}(12-14),  1419--1434 (2021)

\bibitem{sharma2019third}
Sharma, P., Pathak, D., Gupta, A.: Third-person visual imitation learning via decoupled hierarchical controller. Neural Information Processing Systems  (2019)

\bibitem{shridhar2023perceiver}
Shridhar, M., Manuelli, L., Fox, D.: Perceiver-actor: A multi-task transformer for robotic manipulation. In: Conference on Robot Learning (2023)

\bibitem{singer2023makeavideo}
Singer, U., Polyak, A., Hayes, T., Yin, X., An, J., Zhang, S., Hu, Q., Yang, H., Ashual, O., Gafni, O., Parikh, D., Gupta, S., Taigman, Y.: Make-a-video: Text-to-video generation without text-video data. In: International Conference on Learning Representations (2023)

\bibitem{sohn2015learning}
Sohn, K., Lee, H., Yan, X.: Learning structured output representation using deep conditional generative models. Neural Information Processing Systems  (2015)

\bibitem{torabi2018behavioral}
Torabi, F., Warnell, G., Stone, P.: Behavioral cloning from observation. In: International Joint Conference on Artificial Intelligence (2018)

\bibitem{torabi2018generative}
Torabi, F., Warnell, G., Stone, P.: Generative adversarial imitation from observation. arXiv preprint arXiv:1807.06158  (2018)

\bibitem{torabi2019imitation}
Torabi, F., Warnell, G., Stone, P.: Imitation learning from video by leveraging proprioception. In: International Joint Conference on Artificial Intelligence (2019)

\bibitem{trott2019keeping}
Trott, A., Zheng, S., Xiong, C., Socher, R.: Keeping your distance: Solving sparse reward tasks using self-balancing shaped rewards. In: Neural Information Processing Systems (2019)

\bibitem{van2017neural}
Van Den~Oord, A., Vinyals, O., , Kavukcuoglu, K.: Neural discrete representation learning. Neural Information Processing Systems  (2017)

\bibitem{vaswani2017attention}
Vaswani, A., Shazeer, N., Parmar, N., Uszkoreit, J., Jones, L., Gomez, A.N., Kaiser, {\L}., Polosukhin, I.: Attention is all you need. Neural Information Processing Systems  (2017)

\bibitem{villegas2022phenaki}
Villegas, R., Babaeizadeh, M., Kindermans, P.J., Moraldo, H., Zhang, H., Saffar, M.T., Castro, S., Kunze, J., Erhan, D.: Phenaki: Variable length video generation from open domain textual descriptions. In: International Conference on Learning Representations (2023)

\bibitem{villegas2019high}
Villegas, R., Pathak, A., Kannan, H., Erhan, D., Le, Q.V., Lee, H.: High fidelity video prediction with large stochastic recurrent neural networks. In: Neural Information Processing Systems (2019)

\bibitem{walker2021predicting}
Walker, J., Razavi, A., Oord, A.v.d.: Predicting video with vqvae. arXiv preprint arXiv:2103.01950  (2021)

\bibitem{wang2024worlddreamer}
Wang, X., Zhu, Z., Huang, G., Wang, B., Chen, X., Lu, J.: Worlddreamer: Towards general world models for video generation via predicting masked tokens. arXiv preprint arXiv:2401.09985  (2024)

\bibitem{wu2023unleashing}
Wu, H., Jing, Y., Cheang, C., Chen, G., Xu, J., Li, X., Liu, M., Li, H., Kong, T.: Unleashing large-scale video generative pre-training for visual robot manipulation. arXiv preprint arXiv:2312.13139  (2023)

\bibitem{wu2023pre}
Wu, J., Ma, H., Deng, C., Long, M.: Pre-training contextualized world models with in-the-wild videos for reinforcement learning. arXiv preprint arXiv:2305.18499  (2023)

\bibitem{wu2018laplacian}
Wu, Y., Tucker, G., Nachum, O.: The laplacian in rl: Learning representations with efficient approximations. In: International Conference on Learning Representations (2018)

\bibitem{xiao2022masked}
Xiao, T., Radosavovic, I., Darrell, T., Malik, J.: Masked visual pre-training for motor control. arXiv preprint arXiv:2203.06173  (2022)

\bibitem{xu2020spatial}
Xu, M., Dai, W., Liu, C., Gao, X., Lin, W., Qi, G.J., Xiong, H.: Spatial-temporal transformer networks for traffic flow forecasting. arXiv preprint arXiv:2001.02908  (2020)

\bibitem{yan2023temporally}
Yan, W., Hafner, D., James, S., Abbeel, P.: Temporally consistent transformers for video generation. In: International Conference on Machine Learning (2023)

\bibitem{yan2021videogpt}
Yan, W., Zhang, Y., Abbeel, P., Srinivas, A.: Videogpt: Video generation using vq-vae and transformers. arXiv preprint arXiv:2104.10157  (2021)

\bibitem{yang2019imitation}
Yang, C., Ma, X., Huang, W., Sun, F., Liu, H., Huang, J., Gan, C.: Imitation learning from observations by minimizing inverse dynamics disagreement. Neural Information Processing Systems  (2019)

\bibitem{yang2024spatiotemporal}
Yang, J., Liu, B., Fu, J., Pan, B., Wu, G., Wang, L.: Spatiotemporal predictive pre-training for robotic motor control. arXiv preprint arXiv:2403.05304  (2024)

\bibitem{yang2021representation}
Yang, M., Nachum, O.: Representation matters: Offline pretraining for sequential decision making. In: International Conference on Machine Learning (2021)

\bibitem{ye2022become}
Ye, W., Zhang, Y., Abbeel, P., Gao, Y.: Become a proficient player with limited data through watching pure videos. In: International Conference on Learning Representations (2022)

\bibitem{yu2023magvit}
Yu, L., Cheng, Y., Sohn, K., Lezama, J., Zhang, H., Chang, H., Hauptmann, A.G., Yang, M.H., Hao, Y., Essa, I., et~al.: Magvit: Masked generative video transformer. In: Computer Vision and Pattern Recognition (2023)

\bibitem{yu2022leverage}
Yu, T., Kumar, A., Chebotar, Y., Hausman, K., Finn, C., Levine, S.: How to leverage unlabeled data in offline reinforcement learning. In: International Conference on Machine Learning (2022)

\bibitem{yu2020meta}
Yu, T., Quillen, D., He, Z., Julian, R., Hausman, K., Finn, C., Levine, S.: Meta-world: A benchmark and evaluation for multi-task and meta reinforcement learning. In: Conference on Robot Learning (2020)

\bibitem{yu2020intrinsic}
Yu, X., Lyu, Y., Tsang, I.: Intrinsic reward driven imitation learning via generative model. In: International Conference on Machine Learning (2020)

\bibitem{zhang2022learning}
Zhang, Q., Peng, Z., Zhou, B.: Learning to drive by watching youtube videos: Action-conditioned contrastive policy pretraining. In: European Conference on Computer Vision (2022)

\bibitem{zheng2023semi}
Zheng, Q., Henaff, M., Amos, B., Grover, A.: Semi-supervised offline reinforcement learning with action-free trajectories. In: International Conference on Machine Learning (2023)

\bibitem{zhou2023learning}
Zhou, B., Li, K., Jiang, J., Lu, Z.: Learning from visual observation via offline pretrained state-to-go transformer. Neural Information Processing Systems  (2023)

\end{thebibliography}
